\documentclass[conference]{IEEEtran}

\usepackage{cite}
\usepackage{amsmath,amssymb,amsfonts}
\usepackage{algorithmic}
\usepackage{graphicx}
\usepackage{textcomp}
\usepackage{xcolor}

\usepackage{mathrsfs}
\usepackage{optidef} 
\usepackage{bm}
\usepackage{subcaption}
\usepackage{stackengine}
\usepackage{tabularray} 
\usepackage{ninecolors}

\definecolor{myblue}{RGB}{0, 128, 200}
\definecolor{mycyan}{RGB}{15, 201, 255}
\definecolor{myyellow}{RGB}{255, 200, 59}

\usepackage[ruled,linesnumbered]{algorithm2e}
\usepackage{setspace}

\DeclareMathAlphabet{\mathpzc}{OT1}{pzc}{m}{it}

\newcommand{\Pop}{\mathpzc{P}}
\newcommand{\A}{\mathpzc{A}}

\newcommand{\algon}[1]{{\small \textsf{#1}}} 
\def\BibTeX{{\rm B\kern-.05em{\sc i\kern-.025em b}\kern-.08em
    T\kern-.1667em\lower.7ex\hbox{E}\kern-.125emX}}

\begin{document}

\title{A Hybrid Evolutionary Approach for Multi Robot Coordinated Planning at Intersections}

\author{\IEEEauthorblockN{Victor Parque}
\IEEEauthorblockA{\textit{Graduate School of Advanced Science and Engineering} \\
\textit{Hiroshima University}\\
1-4-1 Kagamiyama, Higashi-Hiroshima, Hiroshima Prefecture, 739-8527, Japan \\
parque@hiroshima-u.ac.jp, https://orcid.org/0000-0001-7329-1468}
}

\maketitle

\begin{abstract}

Coordinated multi-robot motion planning at intersections is key for safe mobility in roads, factories and warehouses. The rapidly exploring random tree (RRT) algorithms are popular in multi-robot motion planning. However, generating the graph configuration space and searching in the composite tensor configuration space is computationally expensive for large number of sample points. In this paper, we propose a new evolutionary-based algorithm using a parametric lattice-based configuration and the discrete-based RRT for collision-free multi-robot planning at intersections. Our computational experiments using complex planning intersection scenarios have shown the feasibility and the superiority of the proposed algorithm compared to seven other related approaches. Our results offer new sampling and representation mechanisms to render optimization-based approaches for multi-robot navigation.

\end{abstract}


\section{Introduction}

The motion planning of multi-robots in constrained navigation environments is crucial to enable the safe and smooth navigation in lanes and warehouses. Along with recent developments in autonomous navigation and self-driving systems, the motion planning algorithms and scheduling platforms for constrained and intersection environments are part in the agenda for safer and sustainable transportation

The planning and scheduling algorithms for unregulated intersections is an NP-hard problem\cite{inter16}, yet approximations using mixed-integer linear programs and Model Predictive Control (MPC) exist\cite{knei20}. Also, trajectory negotiation in multigraphs\cite{knei19}, the rule-based control\cite{gua14}, as well discrete-based mechanisms exist\cite{sayin19}. Approaches for motion planning at intersections are rule-based, optimization-based, and machine learning-based. A review on the topic can be found at\cite{int19}. 

Within the class of optimization-based approaches, branch and bound\cite{peng17}, genetic algorithm\cite{lin17}, mixed integer linear programs\cite{fay18}, ant colony\cite{wu12}, dynamic programming\cite{fei09,liu19}, mixed-integer quadratic programming\cite{alt16}, mixed-integer non-linear programming\cite{mir19}, model predictive control\cite{aaron24}, conflict-based search\cite{akma24} and the multi-objective evolutionary strategies\cite{ripon17} are often the most studied in multi-robot motion planning.

Multi-robot motion planning is computationally challenging, in which centralized/coordinated algorithms have emerged recently. For instance, the coordinated planner \cite{sve98} uses an explicit multi-robot search space, \cite{guo02} uses D* to split the multi-robot planning and to allow coordination among robots, and the M* plans in the joint configuration space when robots interact\cite{wagner11}; and the discrete version of the rapidly exploring random trees (RRT) builds roadmaps for each robot and then searches in composite configuration space (tensor product of the graph)\cite{solo16}. The approach in \cite{jur09} decouples the multi-robot planning into sequential plans.

The class of RRT-based algorithms are advantageous due to the sampling-based mechanisms that build composite motions. And dRRT*, the asymptotically optimal version of dRRT, allows to rewire the tree to refine paths by pruning the solutions by branch and bound, and by expanding the composite tree heuristically\cite{rahul20}. The fast-dRRT is the computationally efficient version of dRRT*, it resolves the multi-robot collisions in confined environments by allowing some robots to navigate and others to wait\cite{clay20}. Instead of building a single probabilistic roadmap\cite{kavra96}, \cite{irc23} constructed a parametric roadmap and gradient-free heuristic searched over optimal multi-robot trajectory configurations.

Among the above-mentioned, the class of optimization-based approaches are attractive for their ability to solve practical scenarios. In this paper, we propose a new gradient-free optimization scheme to tackle the multi-robot coordinated planning at intersections, and evaluate its performance. The proposed algorithm implements the difference of vectors and the rank-archive mechanisms to better guide the sampling of road-maps useful to render collision-free multi-robot trajectories. The proposed algorithm is coined as the Rank-based Differential Evolution with a Successful Archive (\algon{RADES}). Our contribution is as follows:

\begin{itemize}

\item We overview the class of multi-robot trajectory planning using the lattice-based road-map configurations.

\item We propose the Rank-based Differential Evolution with a Successful Archives (\algon{RADES}) that implements the rank-based selection mechanisms, the archives of successful mutations, and the stagnation control to guide the sampling mechanisms.

\item Our computational experiments involving the multi-robot coordinated planning on ten types of intersection scenarios have shown the feasibility and the superiority of the proposed algorithm compared to seven other related optimization approaches.

\end{itemize}

\begin{figure}[t]
    \centering
    \includegraphics[width=0.98\linewidth]{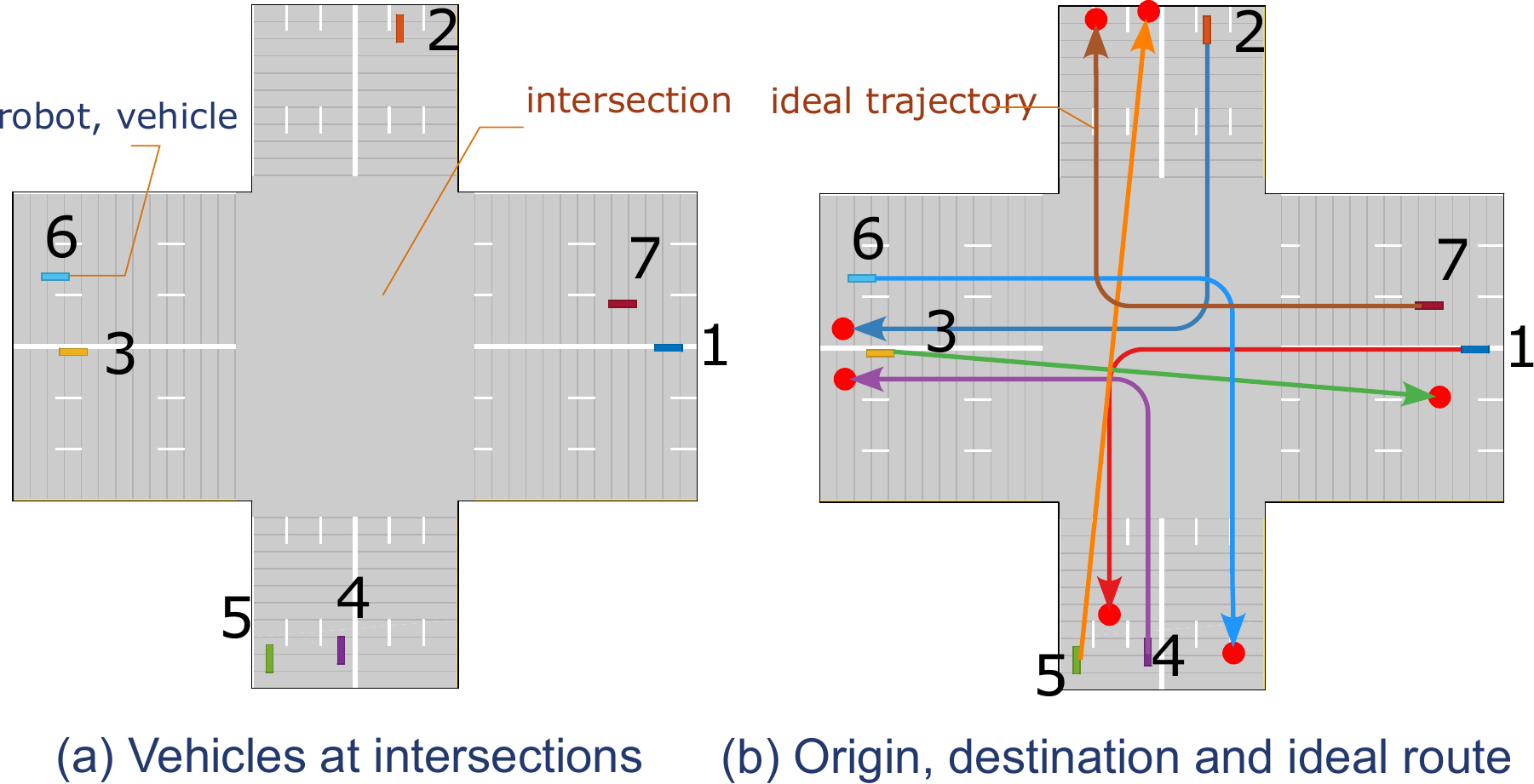}
    \caption{Overview of a cross intersection scenario with 7 robots (vehicles): (a) initial configuration of robots, and (b) the origin (with numbers), the destination (\textcolor{red}{red color}) and the ideal trajectory configurations with arrows.}
    \label{intro}
\end{figure}

\section{Proposed Approach}

\subsection{Preliminaries}

This paper tackles the multi-robot coordinated planning problem at intersections, as shown by Fig. \ref{intro}. The input is the initial configuration of the robots and their corresponding destinations as shown by Fig. \ref{intro}-(a, b). Thus, the goal is to compute collision-free trajectories for a team of $N$ robots (vehicles) in intersection domains. Here,
\begin{itemize}
  \item the robot configuration is represented by the tuple $q \in (x, y, \theta, \kappa)^T$, with position in the Euclidean plane at $(x,y)$, orientation $\theta$, and following the dynamics $(\dot{x}, \dot{y}, \dot{\theta}, \dot{\kappa} ) = (\cos \theta, \sin \theta, \kappa, \sigma)$\cite{clay20}.
  \item $q_i$ is the configuration of the $i$-th robot ($i \in [N]$), considering a team of $N$ robots; in which $q^{o}_i$ and $q^{e}_i$ are the initial and end configurations of the $i$-th robot.
  \item $Q = \{q_1, q_2, q_3, ..., q_N\}$ is the set of $N$ robot configurations of the set of $N$ robots.
\end{itemize}

The ideal/reference local trajectories can be computed by feasible shortest-paths over the intersection domains as shown by Fig. \ref{intro}-(b).

\subsection{Multi-Robot Coordinated Planning}

We tackle the multi-robot coordinated planing problem by solving the following:

\begin{equation}\label{opti}
\begin{aligned}
& \underset{x}{\text{Minimize}}
& & \sum_{i=1}^{N} L(P_i) \\
& \text{subject to}
& & P_i \subset E_i, \; i = 1, \ldots, N.
\end{aligned}
\end{equation}
where

\begin{itemize}

    \item $P_i = (e^1_i, e^2_i, ..., e^k_i,..., e^{n-1}_i)$ represents the sequence of multi-robot collision-free edges rendered from a roadmap graph $G_i = (V_i,~E_i)$ and computed by dRRT*\cite{rahul20,clay20}. 

    \item $L(P_i)$ denotes the length of the trajectory $P_i$ (Euclidean norm), as the result of the sum of lengths of edges.
\end{itemize}

The basic idea of the above is to search over the space of the composite roadmap configurations to guide dRRT* to navigate and sample points in the constrained search space. In this paper, each roadmap is constructed using triangular lattice paths defined by base $b_i$ and height $h_i$; thus, it becomes straightforward to search over the space of lattice path configurations by sampling over the space of $b_i$ and $h_i$. The cost function in (\ref{opti}) is nondifferentiable, thus the class of gradient-free optimization algorithms are suitable to tackle the problem.

\begin{figure}[t]
    \centering
    \includegraphics[width=0.98\linewidth]{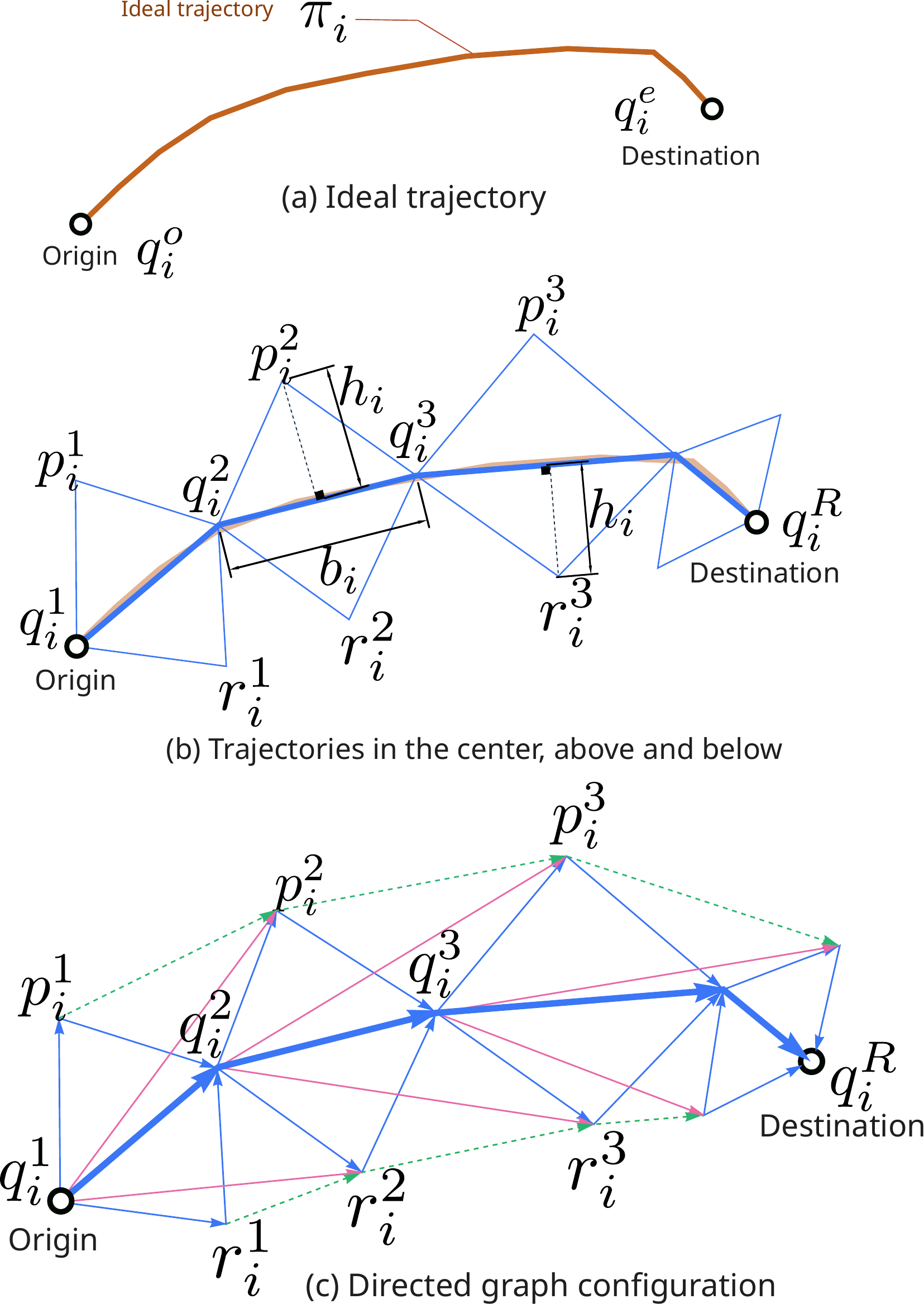}
    \caption{Basic approach for lattice-based roadmap configuration.}
    \label{basic}
\end{figure}

\subsection{Roadmap Lattice Configuration}

We construct the reference trajectories over a lattice-based roadmaps as follows:

\begin{itemize}
    \item An ideal trajectory $\pi_i$ is sampled in the free space to allow navigation from initial configuration $q^{o}_i$ towards the end configuration $q^{e}_i$, as shown by Fig. \ref{basic}-(a). The sampling of the trajectory is performed via feasible shortest navigation paths.
        
    \item Let the set of trajectories $\pi^m_i = \{ q^1_i,~ q^2_i,~ q^3_i, ..., ~q^R_i\}$, $\pi^a_i = \{ p^1_i,~ p^2_i,~ p^3_i, ...\}$ and $\pi^b_i = \{ r^1_i, ~r^2_i, ~r^3_i, ...\}$ be the trajectory configurations \emph{on}, \emph{above} and \emph{below} the ideal trajectory $\pi_i$ computed by discretization with triangular elements with base $b_i$ and height $h_i$ as shown by Fig. \ref{basic}-(b). Here, $q^1_i = q^{o}_i$ and $q^R_i = q^{e}_i$, and

        \begin{equation}\label{eqR}
          R = 1+ \displaystyle \frac{L(\pi_i)}{b_i}
        \end{equation}
        is the number of points in the set $\pi^m_i$, and $L(\pi_i)$ is the length of the trajectory $\pi_i$.
    
    \item We generate the roadmap for the $i$-th robot by constructing the directed graph $G_i = (V_i,~E_i)$ where the set of nodes is constructed by
        \begin{equation}\label{Vi}
          V_i = \pi^m_i \cup \pi^a_i \cup \pi^b_i,
        \end{equation}
        and the set of edges $E_i$ is constructed with        

\end{itemize}

\begin{multline}
    E_i = \Big \{ e \in E |~ e = (q^{j}_i,~ q^{j+1}_i) \vee e = (q^{j}_i, ~p^{j}_i) \vee e = (q^{j}_i, ~r^{j}_i) \vee \\ e = (p^{j}_i, ~q^{j+1}_i) \vee e = (r^{j}_i, ~q^{j+1}_i)  \vee e = (p^{j}_i, ~p^{j+1}_i)  \vee \\ e = (r^{j}_i, ~r^{j+1}_i) \vee e = (q^{j}_i, ~p^{j+1}_i) \vee e = (q^{j}_i, ~r^{j+1}_i) \Big \}
\end{multline}

In the above, edges in the graph $G_i$ are directed. Fig. \ref{basic}-(c) shows an example of the roadmap construction. The basic idea of the above is to construct a feasible and navigable lattice path for each robot based on a directed graph configurations and the set of trajectories $\pi^m_i$, $\pi^a_i$, and $\pi^b_i$, $i \in [N]$. Furthermore, the ideal/reference trajectory $\pi_i$ can be rendered through known single shortest and feasible path algorithms over the free space; and the trajectories $\pi^m_i, \pi^a_i$ and $\pi^b_i$ are discretizations based on the ideal trajectory $\pi_i$. The constructed directed graph $G_i$ allows the $i$-th robot to navigate close to the ideal trajectory while allowing a margin of safety/deviation based on parameters $b_i$ and $h_i$. 

The set $P_i$ denotes the trajectory transition in the free navigable space. Here, let $(v^1_i, v^2_i, ..., v^k_i,..., v^n_i)$ be the sequence of vertices of the trajectory ($P_i \subset V_i$) such that $e^k_i = (v^k_i, v^{k+1}_i)$ is the edge incident to $v^k_i$ and $v^{k+1}_i$, and $e^k_i \in E_i,~ v_1 = q^{\circ}_i,~ v^n_i = q^e_i$.

\subsection{Rank-based Differential Evolution with a Successful Archive (\algon{RADES})}

In this paper, we propose \algon{RADES}, a gradient-free optimization algorithm inspired by the difference of vectors in Differential Evolution. Here, solutions are sampled by using the following relations:

\begin{equation}\label{de}
    \bm{x}_{i, t+1} =
  \begin{cases}
    \bm{u}_{i, t}, & \text{for } f(\bm{u}_{i, t}) < f(\bm{x}_{i, t}) \\
    \bm{x}_{i, t}, & \text{otherwise} \\
  \end{cases}
\end{equation}

\begin{equation}\label{ut}
    \bm{u}_{i, t} = \bm{x}^r_{i, t} + \bm{b}_{i, t} \circ (\bm{v}_{i, t} - \bm{x}^r_{i, t})
\end{equation}

\begin{equation}\label{xref}
    \bm{x}^r_{i, t} =
  \begin{cases}
    \bm{x}_{i, t}, & q_i \leq Q\\
    \A_{i,t}, & \text{otherwise} \\
  \end{cases}
\end{equation}

\begin{equation}\label{vt}
    \bm{v}_{i, t} =
  \begin{cases}
    \bm{x}_{ \text{best} } + F (\bm{x}_{r_1} - \bm{x}_{r_2}), & q_i \leq Q \\
    \A_{\text{best}} + F (\A_{r_1} - \A_{r_2}), & \text{otherwise} \\
  \end{cases}
\end{equation}

\begin{equation}\label{r1r2}
  r_{1, 2} = \Bigg \lfloor   \frac{|\Pop|}{2\beta -1 } \Big ( \beta - \sqrt{\beta^2 - 4 (\beta-1) r }\Big ) \Bigg \rfloor
\end{equation}

\begin{equation}
    \bm{b}_{i, t} = ( b_{t,1}, ...,b_{t,k}, ..., b_{t,D} )
\end{equation}

\begin{equation}\label{btk}
    b_{t,k} =
  \begin{cases}
    1, & \text{for } r_{t,k} < CR \text{ or } k = jrand\\
    0, & \text{otherwise} \\
  \end{cases}
\end{equation}

\begin{algorithm}[t]
\setstretch{1.2}

$FEs = 0$, $t = 0$, $q_i = 0$\;

Generate a set of $N$ individuals randomly as initial population set $\Pop$;

Initialize the archive $\A$ from the population set $\Pop$;

Initialize the $F$ and $CR$\;

$FEs = FEs + N$\;

\While{$FEs \leq MaxFEs$}{

    $t = t + 1$\;


    Find out the best individual $\bm{x}_{\text{best}}$ and $\A_{\text{best}}$ from the population and the archive $\A$ respectively\;

    \For{$i=1$ \KwTo $NP$}{
    
        Generate $r_1$ and $r_2$ using (\ref{r1r2})\;
        
        Generate the mutant vector $\bm{v}_{i, t}$ using (\ref{vt})\;
        
        Generate the trial vector $\bm{u}_{i, t}$ using (\ref{ut})\;

        \eIf{$f(\bm{u}_{i, t}) < f(\bm{x}_{i, t})$}
        {
            $\bm{x}_{i, t+1} = \bm{u}_{i, t}$\; 
                        
            $\bm{u}_{i, t} \rightarrow \A$\;
            
            Delete an element from $\A$ if $|\A| > |\Pop|$\;
                        
            $q_i = 0$\;
            
        }{
            $\bm{x}_{i, t+1} = \bm{x}_{i, t}$\;
            
            $q_i = q_i + 1$\;
        }        
        
        $FEs = FEs + 1$\;
    } 
    
}

\caption{\algon{RADES}\label{rades}}
\end{algorithm}

\begin{itemize}
  \item $\bm{x}_t$ is a real vector ($\bm{x}_t \in \mathbb{R}^D$) denoting the configuration of the roadmap $\bm{x} = (b_i, h_i)_{i \in \{1, ..., N \}}$,
  \item $\bm{u}_{t}$ is the trial solution,
  \item $\circ$ is the Hadamard product (element-wise),
  \item $\bm{x}^r_{i, t}$ is the $i$-th reference individual (solution) sampled from either the population $\Pop$ or the archive $\A$, 
  \item $\bm{v}_t$ is the mutant vector,
  \item $\bm{x}_{r_1}$ is the $r_1$-th individual from the sorted population by fitness,
  \item $\A_{r_1}$ is the  $r_1$-th individual of the archive $\A$,
  \item $\bm{b}_t$ is a binary vector,
  \item $r, ~r_{t,k}$ are random numbers in $U[0,1]$,
  \item $jrand$ is a random integer $U[1, D], k \in [1, D],$,
  \item $CR$ is a probability of crossover.
  \item $\beta$ is a bias term,
  \item $q_i$ is the stagnation counter,
  \item $Q$ is the threshold for stagnation counter. 
\end{itemize}

\algon{RADES} evolves a set $\Pop$ of individuals (solutions), whose cardinality is $NP = |\Pop|$, by using initialization and difference of vectors. Algorithm 1 shows the overall mechanism. After the population is initialized, the mutation operator generates the vector $\bm{v}_t$ and the crossover operator generates the trial vector $\bm{u}_t$ (by using Eq. \ref{ut}). A selection operator (Eq. \ref{de}) updates solutions of the population. \algon{RADES} extends previous schemes \cite{de97,rbde,desps,irc23} by incorporating not only the use of an archive $\A$ which has the role of storing successful mutations for subsequent sampling operations, and an stagnation counter $q_i$ which has the role of guiding the source sampling vectors. Furthermore, extending the purely successful parent selection mechanisms\cite{desps}, \algon{RADES} explicitly differentiates the exploitation and exploration in the difference of vectors. When the condition $q_i \leq Q$ is true (false), that is when the search operators lead to improved (stagnated) solutions, both (\ref{xref}) and (\ref{vt}) become exploitative (explorative), thus the suitable exploitation-exploration trade-off is explicitly modeled by choosing a suitable threshold $Q$. \algon{RADES} ends when a user-defined maximum number of function evaluations are met.

Furthermore, the variable $P_i$ in Eq. \ref{opti} can be generated from the lattice-based roadmap configuration $G_i$. Since parameters $b_i$ and $h_i$ define the graph $G_i$, the tuple $\bm{x} = (b_i, h_i)_{i \in \{1, ..., N \}}$ denotes the set of parameters defining the navigation roadmaps (lattices) of the set of $N$ robots.

\section{Computational Experiments}

To evaluate the performance and feasibility of the proposed approach, we conducted computational experiments in the context of multi-robot path planning in intersection situations.

\subsection{Settings}

To allow the modeling of multi-robot planning in distinct scenarios, we use the configurations shown by Fig. \ref{maps}. Here, the origin and destination, as well as the ideal/reference trajectories are given for navigation considering intersections. We used up to 10 robots for the sake of evaluating the feasibility of tackling multi-robot coordinated planning. The map dimensions were set to a squared region and the robot dimensions were set to $3.2 \times 0.8$ units. To tackle Eq. (\ref{opti}), we used \algon{RADES} and compared its performance to other gradient-free population-based Differential Evolution heuristics:

\begin{itemize}
  \item  \algon{BETACODE}: Differential Evolution with Stochastic Opposition-Based Learning and Beta Distribution\cite{betacode16},
  \item \algon{DESIM}: Differential Evolution with Similarity Based Mutation\cite{desim},
  \item \algon{DERAND}: Differential Evolution with DE/rand/1/bin Strategy\cite{de97},
  \item \algon{RBDE}: Rank-based Differential Evolution\cite{rbde}, 
  \item \algon{DCMAEA}: A Differential Covariance Matrix Adaptation Evolutionary Algorithm\cite{dcmaea},
  \item \algon{OBDE}: Opposition-Based Differential Evolution\cite{obde},
  \item \algon{UMS SHADE}: Underestimation-Based Differential Evolution with Success History\cite{ums}
\end{itemize}

The motivation behind using the above set is due to our intention to evaluate related forms of sampling, exploration, exploitation, and selection pressure during optimization. Parameters for optimization included probability of crossover $CR = 0.5$, scaling factor $F = 0.7$, threshold on stagnation $Q= 128$, population size $NP = 10$ for all \algon{DE}-based algorithms, being configurations used in related optimization literature\cite{desps}. The bias term $\beta$ in \algon{RADES} (Eq. \ref{r1r2}) and \algon{RBDE} $\beta = 2$, and the termination criterion $MaxFEs = 300$ function evaluations. The motivation behind using the above parameter set is to allow equal importance for trial solution generation over all search dimensions. Also, due to the stochastic nature of the optimization algorithms, we performed 20 independent runs per intersection navigation instance to evaluate the convergence performance. The bounds for parameter search were set as follows $b_i \in [1, 6]~ h_i \in [0.5, 5]$, which imply the reasonable scope for lattice-based grid map configurations.

\begin{figure*}[t!]
\centering
\begin{tblr}{
    colspec = {Q[l]},
    rowsep = 3pt,
    colsep = 1pt,
    vlines = dashed,
    hlines = dashed,
    }
    
    \stackon{\includegraphics[width=0.33\textwidth]{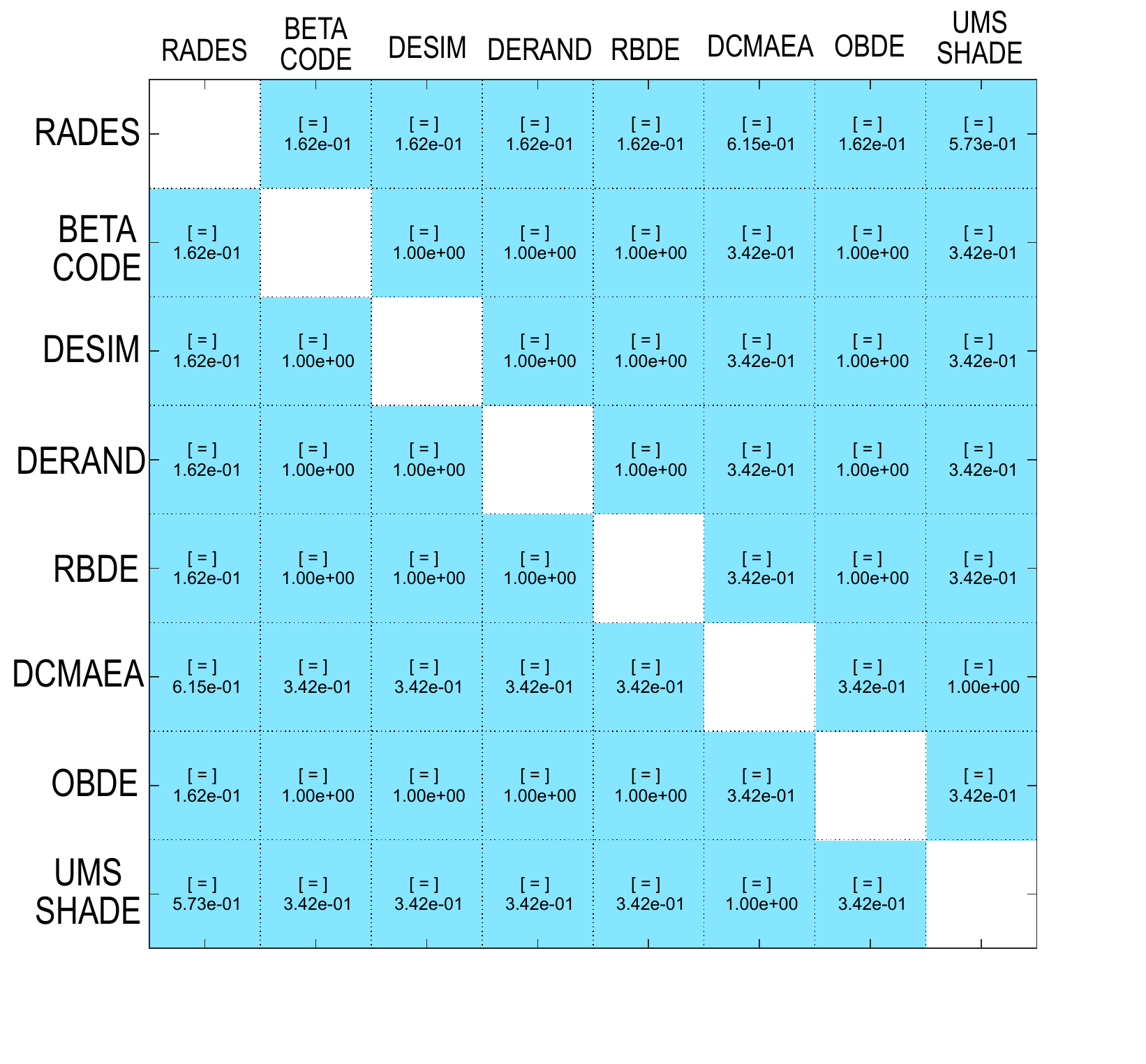}}{\scriptsize Instance 1}
    \stackon{\includegraphics[width=0.33\textwidth]{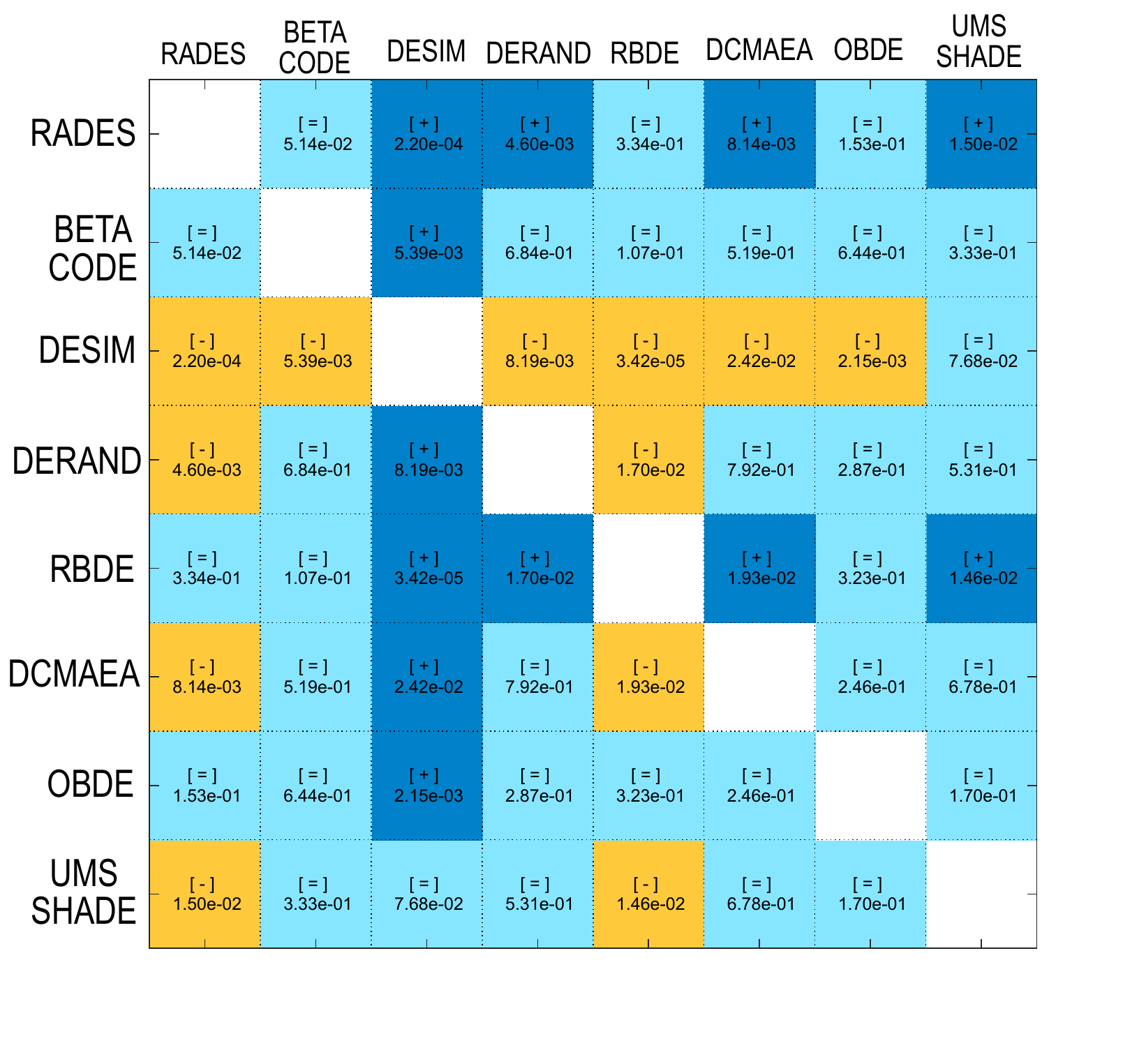}}{\scriptsize Instance 2}
    \stackon{\includegraphics[width=0.33\textwidth]{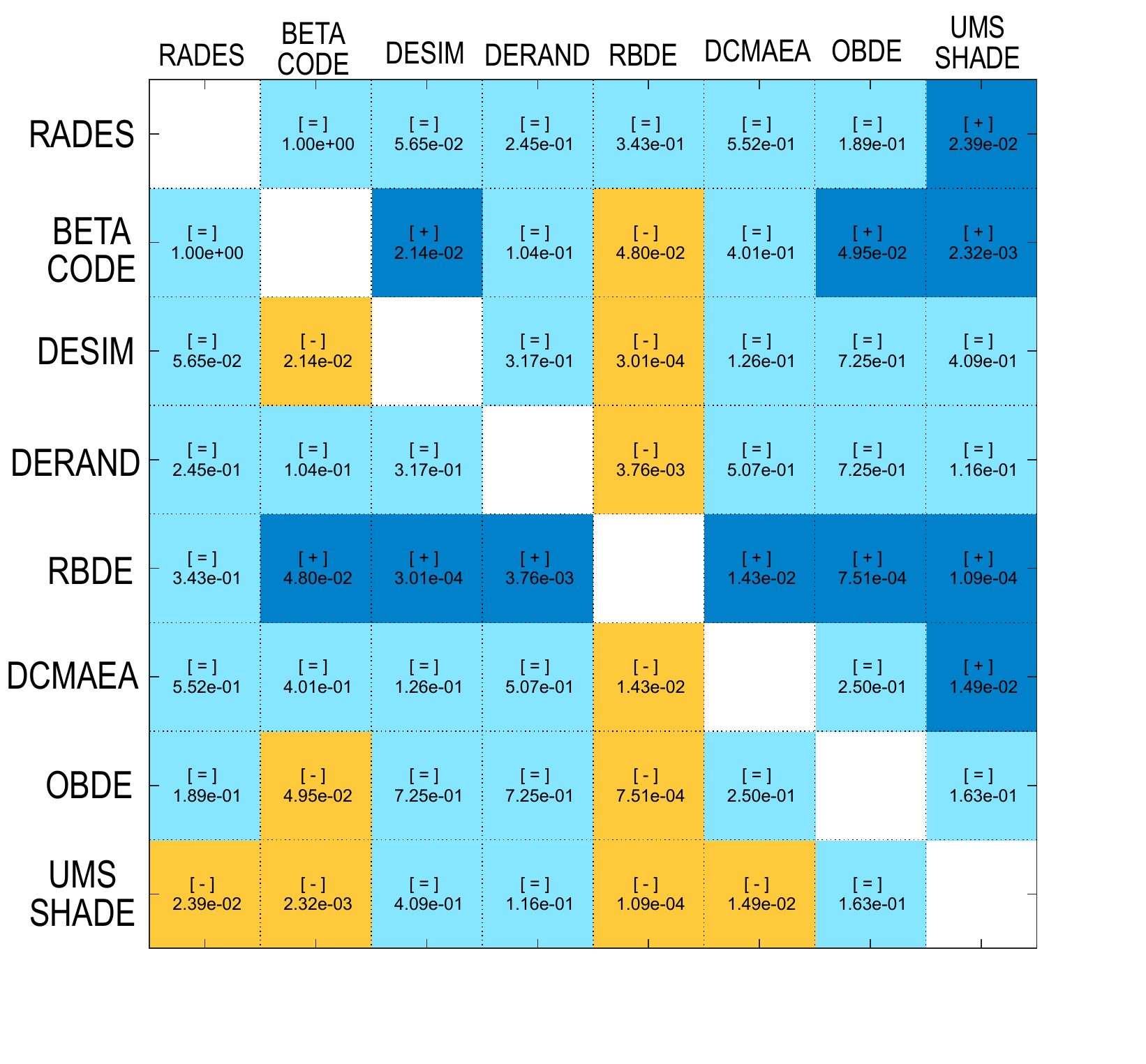}}{\scriptsize Instance 3}
	\\
    \stackon{\includegraphics[width=0.33\textwidth]{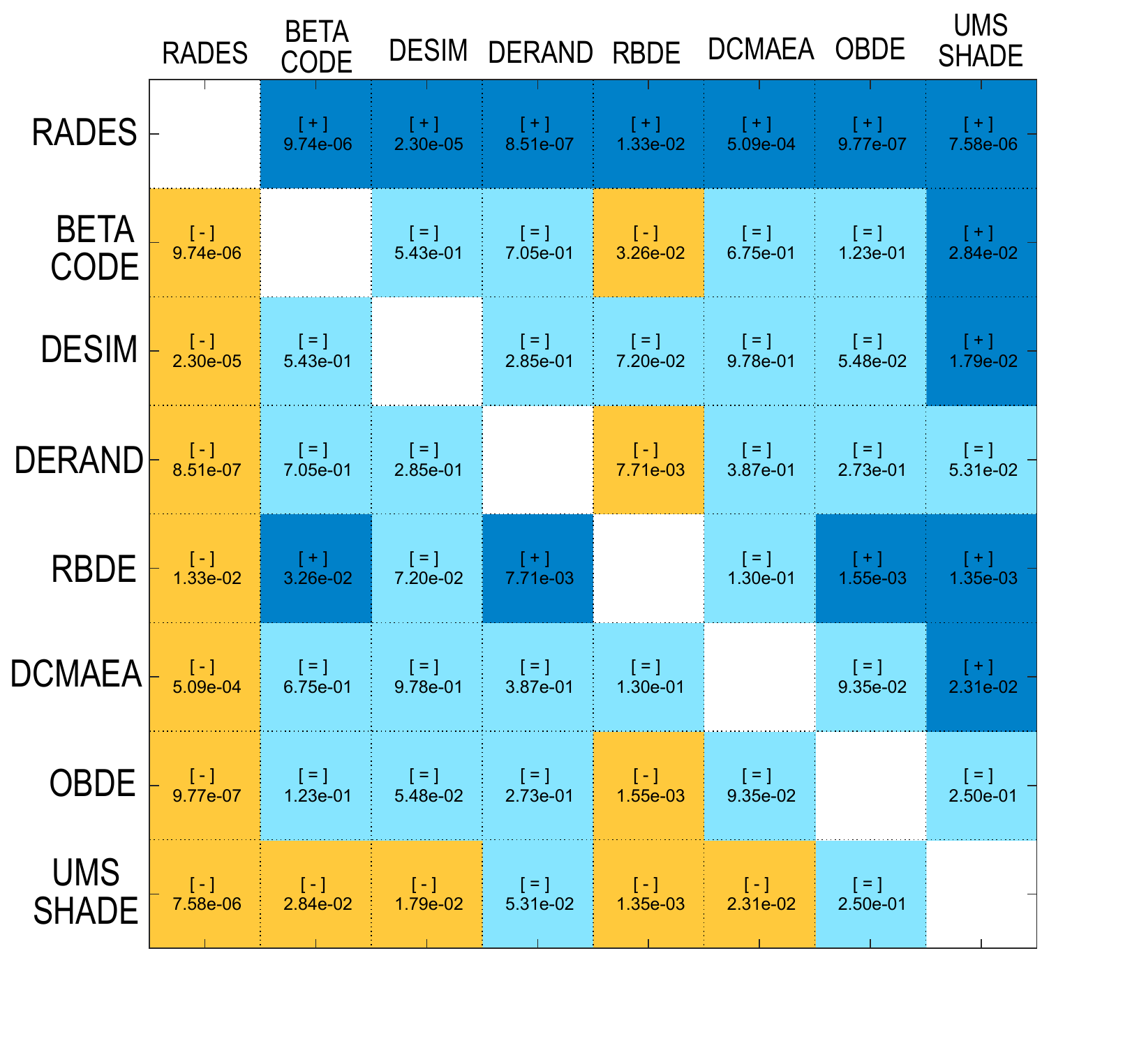}}{\scriptsize Instance 4}
    \stackon{\includegraphics[width=0.33\textwidth]{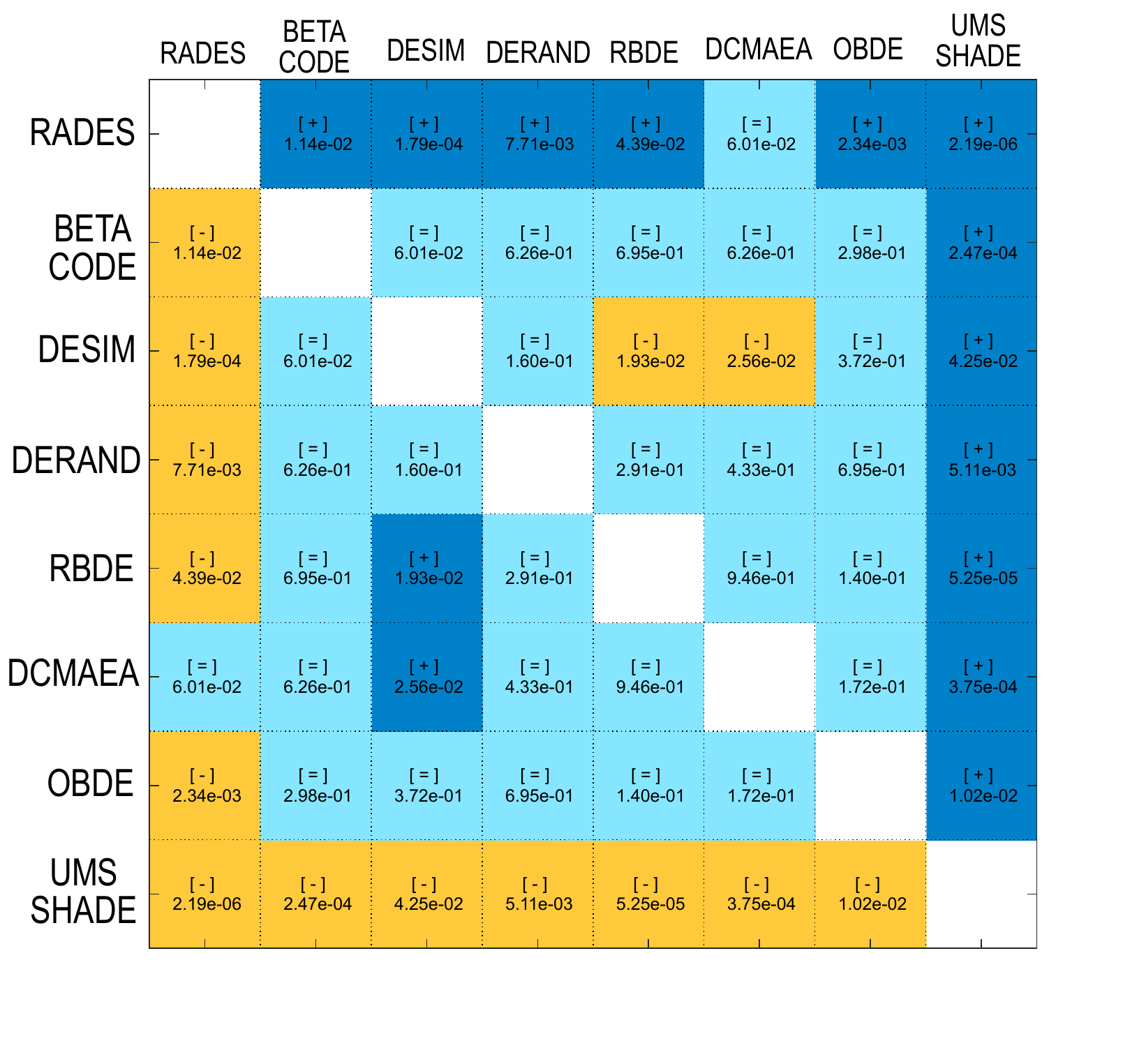}}{\scriptsize Instance 5}
    \stackon{\includegraphics[width=0.33\textwidth]{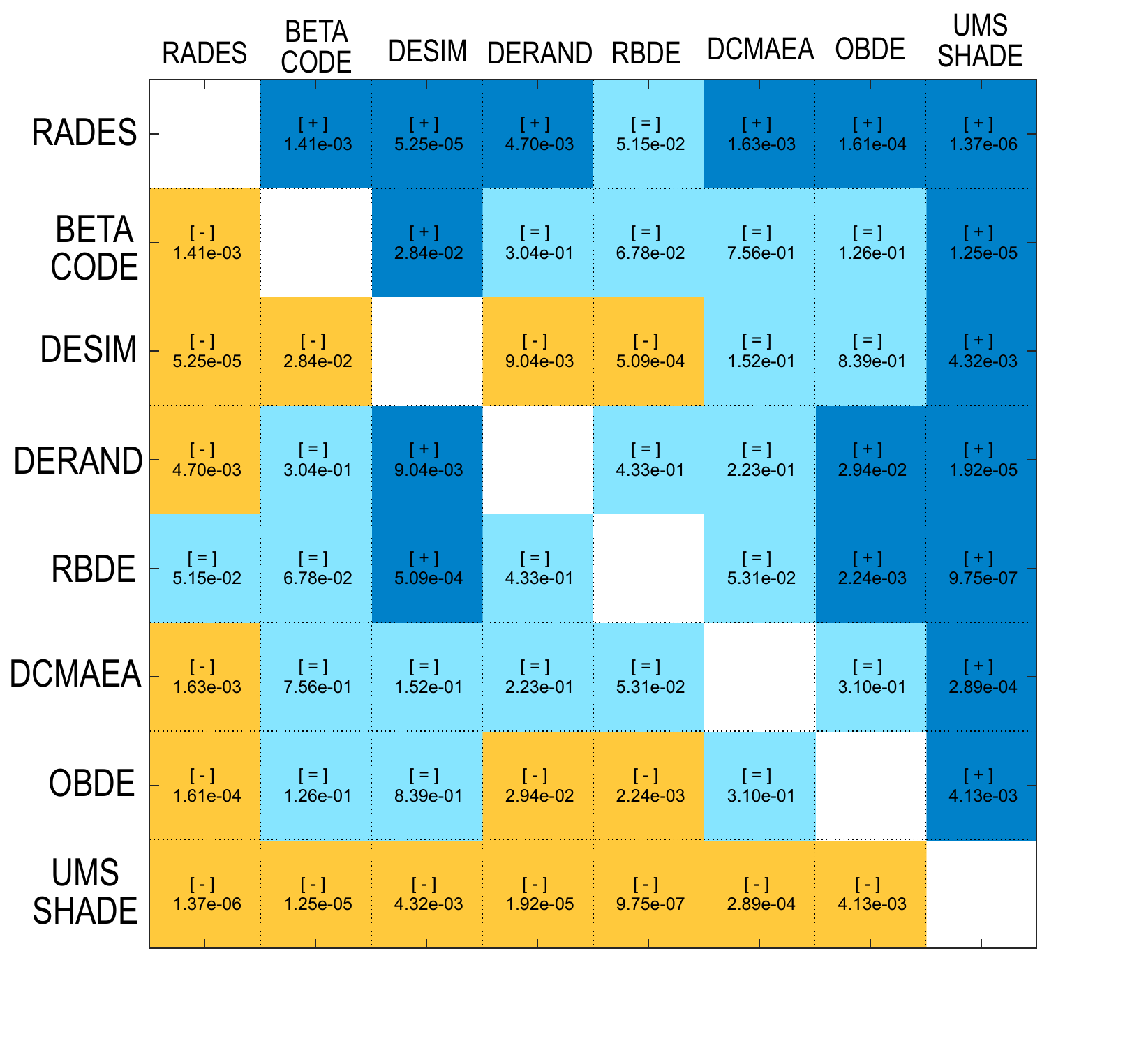}}{\scriptsize Instance 6}
    \\
    \stackon{\includegraphics[width=0.33\textwidth]{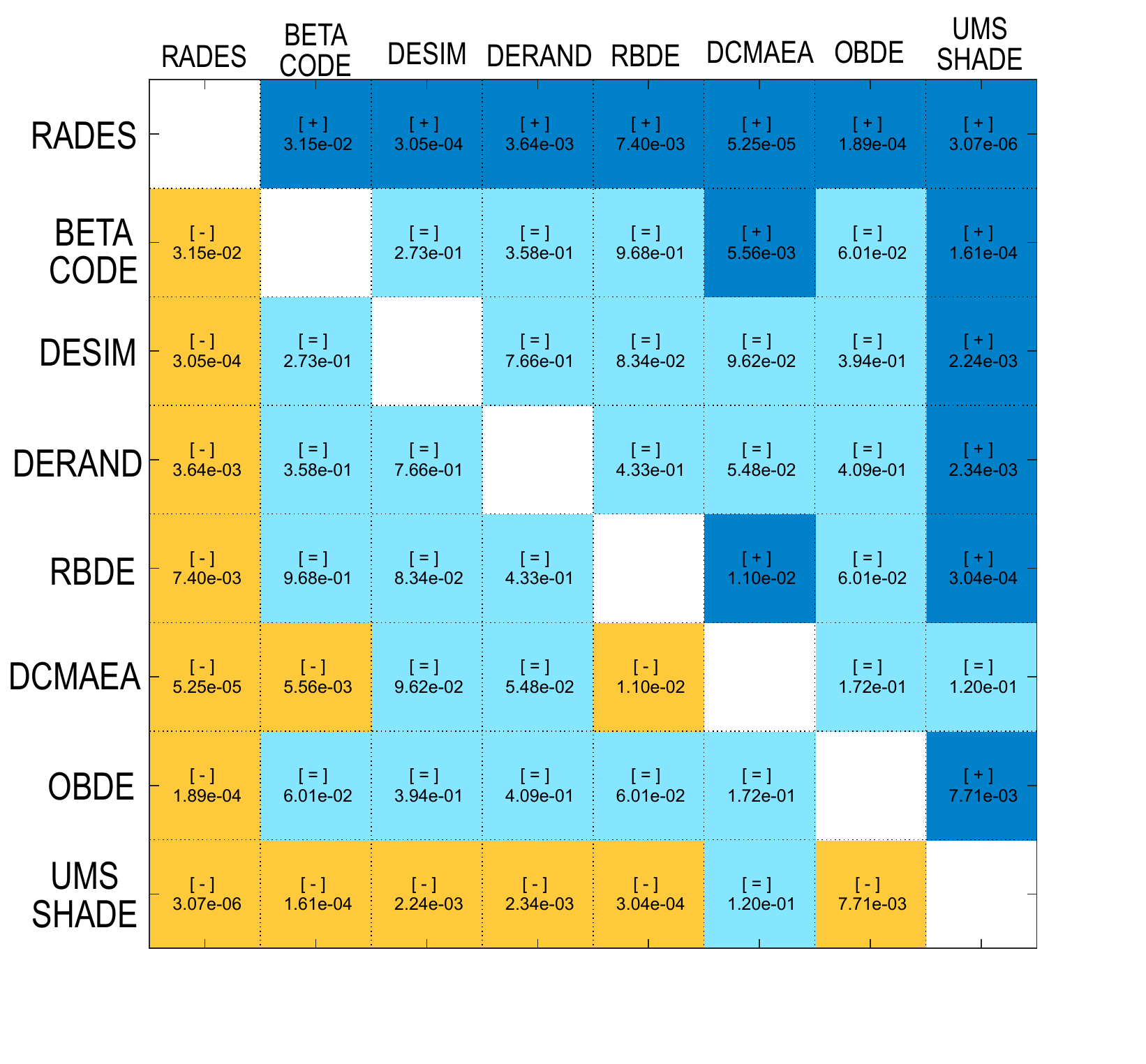}}{\scriptsize Instance 7}
    \stackon{\includegraphics[width=0.33\textwidth]{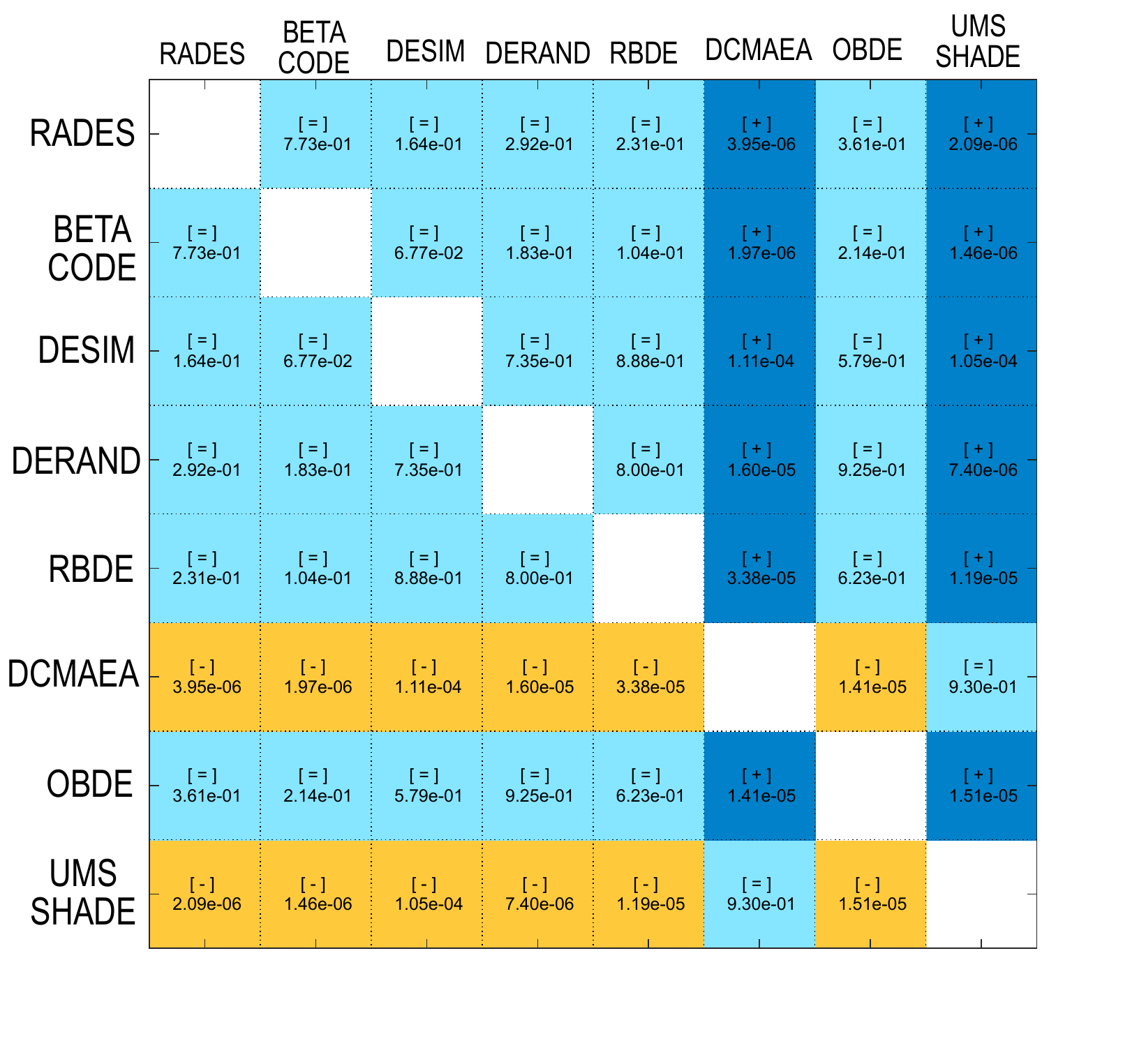}}{\scriptsize Instance 8}
    \stackon{\includegraphics[width=0.33\textwidth]{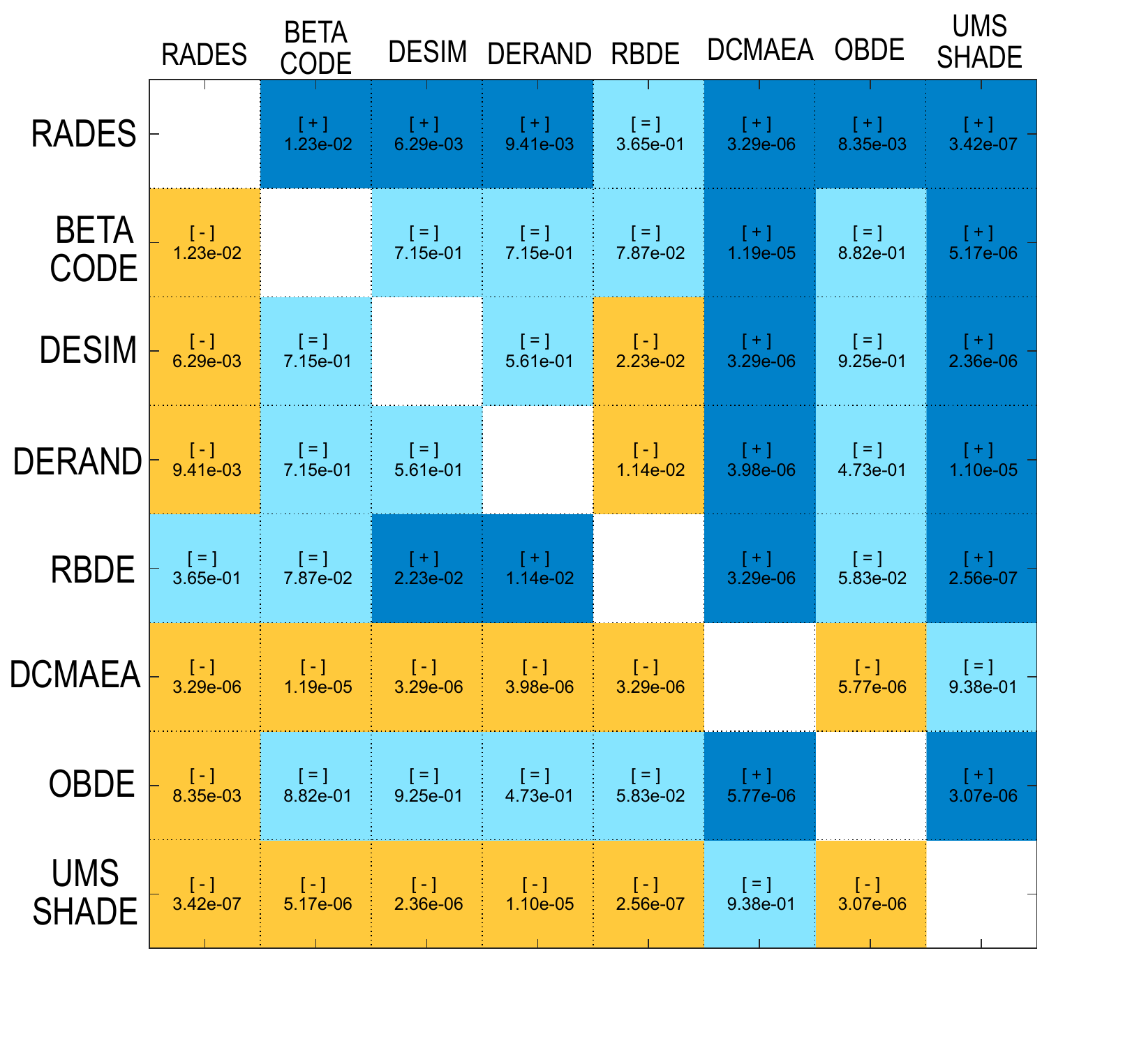}}{\scriptsize Instance 9}
    \\
\end{tblr}
\caption{Statistical comparisons derived from pair-wise Wilcoxon tests at 5\% significance level.}
\label{wil}
\end{figure*}

\begin{figure*}[ht]
    \centering
    \includegraphics[width=0.998\textwidth]{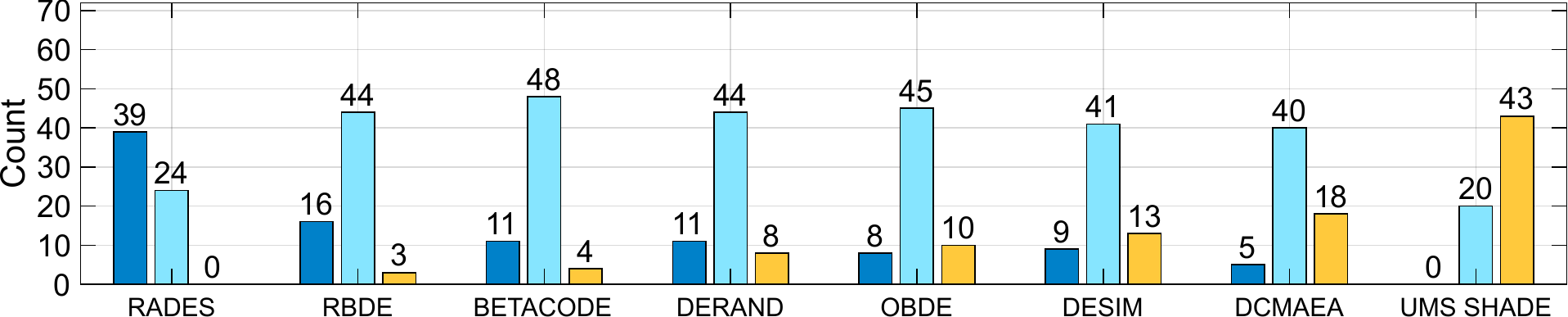}
    \caption{Number of cases of \textcolor{myblue}{ {\small {\textsf{outperformance in blue color}}}}, \textcolor{mycyan}{ {\small {\textsf{equal performance in cyan color}}}}, and \textcolor{myyellow}{ {\small {\textsf{underperformance in yellow color}}}} derived from pair-wise Wilcoxon tests at 5\% significance level. The bars in blue/cyan/orange color show the number of times in which the algorithm is significantly better/similar/worse compared to other algorithms. }
    \label{pv}
\end{figure*}

\subsection{Results and Discussion}

\begin{figure*}[t!]
\centering
\begin{tblr}{
    colspec = {Q[l]},
    rowsep = 3pt,
    colsep = 1pt,
    vlines = dashed,
    hlines = dashed,
    }
    \stackon{\includegraphics[width=0.45\textwidth]{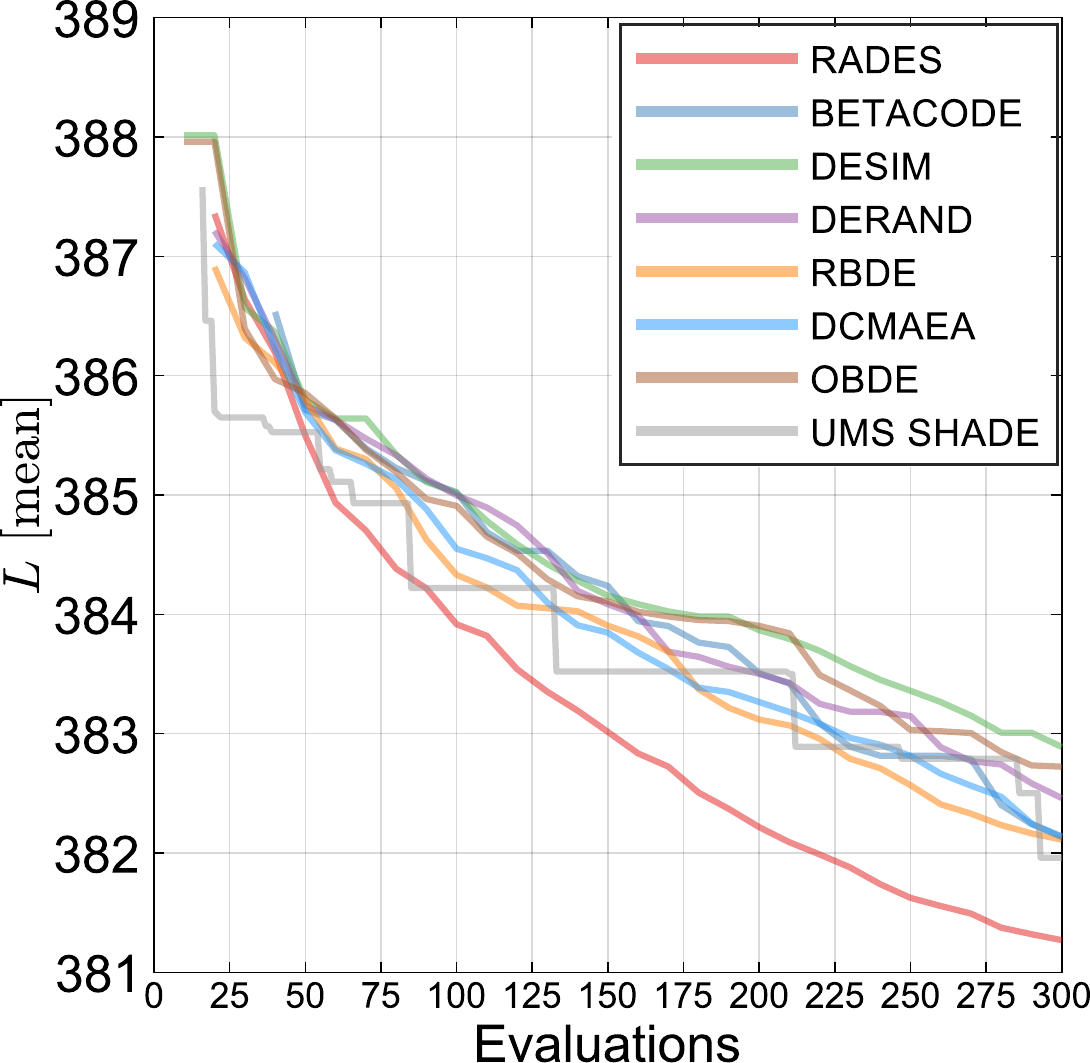}}{\scriptsize Example 1}
    \stackon{\includegraphics[width=0.45\textwidth]{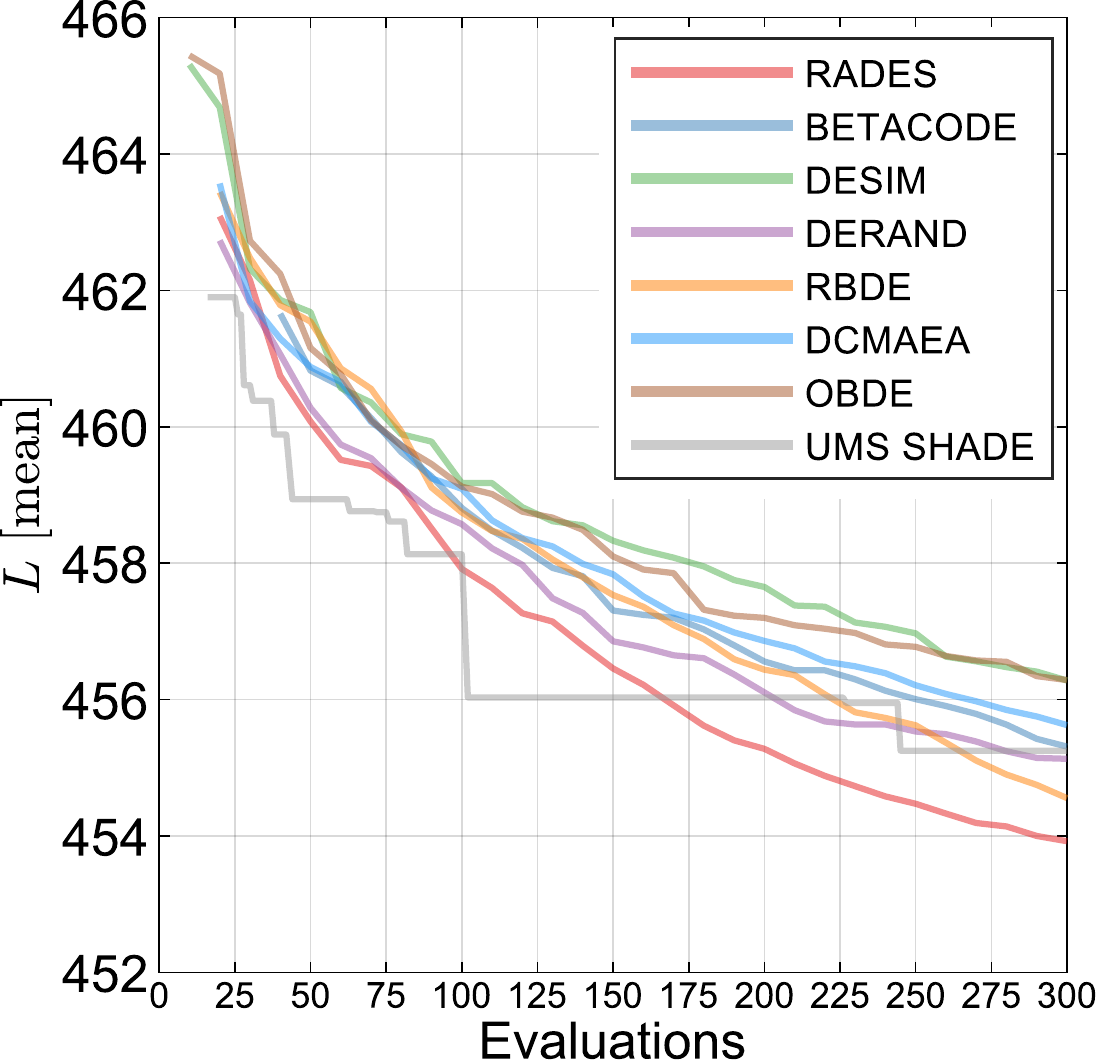}}{\scriptsize Example 2}
	\\
\end{tblr}
\caption{Examples of the average convergence performance over independent runs.}
\label{conv}
\end{figure*}

In order to evaluate the convergence performance among the evaluated algorithms, we conducted a pair-wise statistical comparison based on Wilcoxon test at 5\% significance level. Fig. \ref{wil} shows the pair-wise statistical comparison, in which an algorithm in the row shows whether it is significantly better (+), worse (-) or similar (=) compared to an algorithm in the column. The pairwise comparisons are reported per each navigation instance. And Fig. \ref{pv} summarizes the overall performance across independent runs and navigation instances. Fig \ref{pv} shows the number of times in which the algorithm performs better (bars in \textcolor{myblue}{ {\small {\textsf{blue}}}}), similarly to (bars in \textcolor{mycyan}{ {\small {\textsf{cyan}}}}), or worse than (bars in \textcolor{myyellow}{ {\small {\textsf{yellow}}}}) other algorithms across all evaluated instances. As such, the x-axis of Fig. \ref{pv} shows the algorithm instance, ordered by rank from the best (left) to the worst (right), and the y-axis shows the count of the number of instances. By observing the results in Fig. \ref{pv}, we can note that \algon{RADES} shows the competitive performance in terms of the number of times it outperforms other algorithms followed by \algon{RBDE} and \algon{BETACODE}. On the other hand, algorithms based on exploration and diversity enhancing schemes such as \algon{DERAND}, \algon{DESIM} and \algon{UMS SHADE} underperform overall navigation scenarios. This observation pinpoints towards the merits of using archives and ranks in gradient-free optimization for multi-robot planning problems. To show an example of the convergence performance, Fig. \ref{conv} shows an example of the convergence behaviour of the cost function as a function of the number of function evaluations of the evaluated algorithms. By observing Fig. \ref{conv}, we can note that \algon{RADES} shows the better convergence in comparison with the related algorithms.

\begin{figure}[t!]
\centering
\begin{tblr}{
    colspec = {Q[l]},
    rowsep = 1pt,
    colsep = 1pt,
    vlines = dashed,
    hlines = dashed,
    }
    \stackon{\includegraphics[width=0.325\columnwidth]{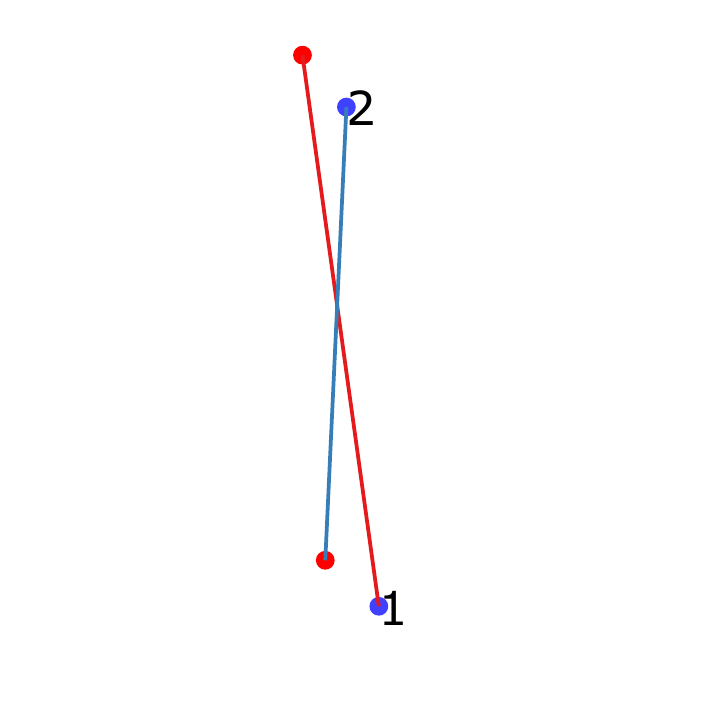}}{\scriptsize Instance 1, 2 robots}
    \stackon{\includegraphics[width=0.325\columnwidth]{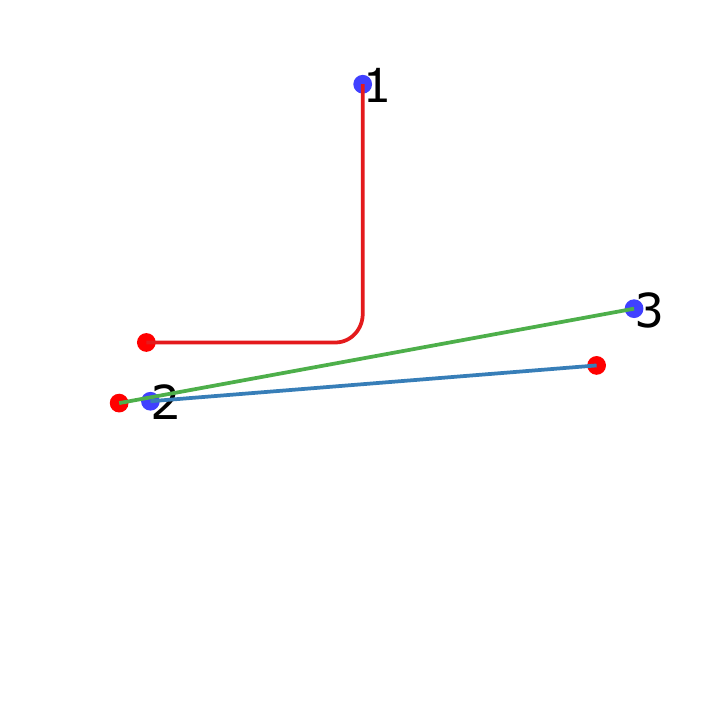}}{\scriptsize Instance 2, 3 robots}
    \stackon{\includegraphics[width=0.325\columnwidth]{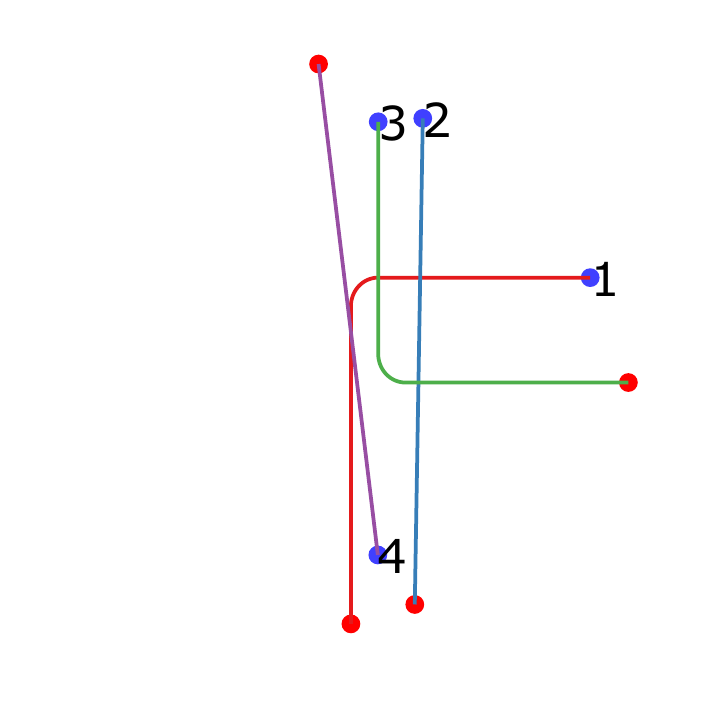}}{\scriptsize Instance 3, 4 robots}
	\\
    \stackon{\includegraphics[width=0.325\columnwidth]{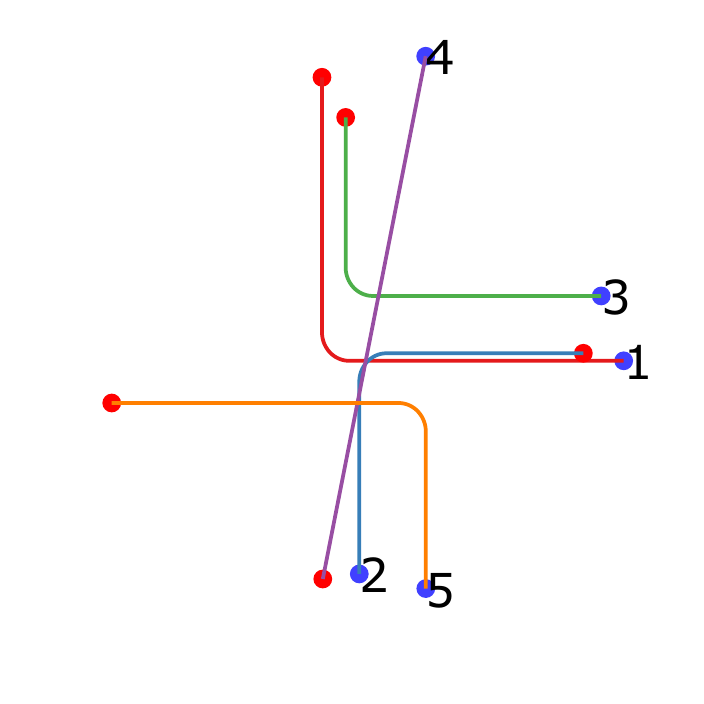}}{\scriptsize Instance 4, 5 robots}
    \stackon{\includegraphics[width=0.325\columnwidth]{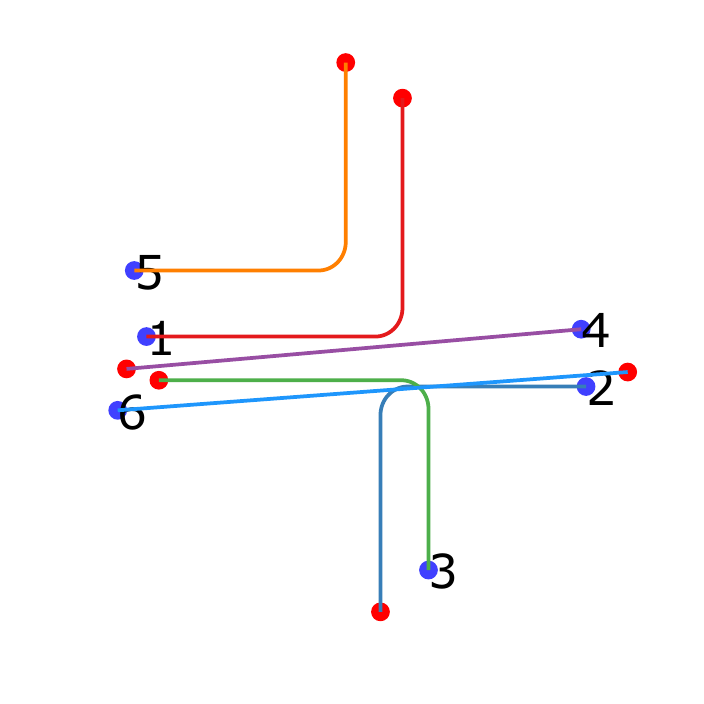}}{\scriptsize Instance 5, 6 robots}
    \stackon{\includegraphics[width=0.325\columnwidth]{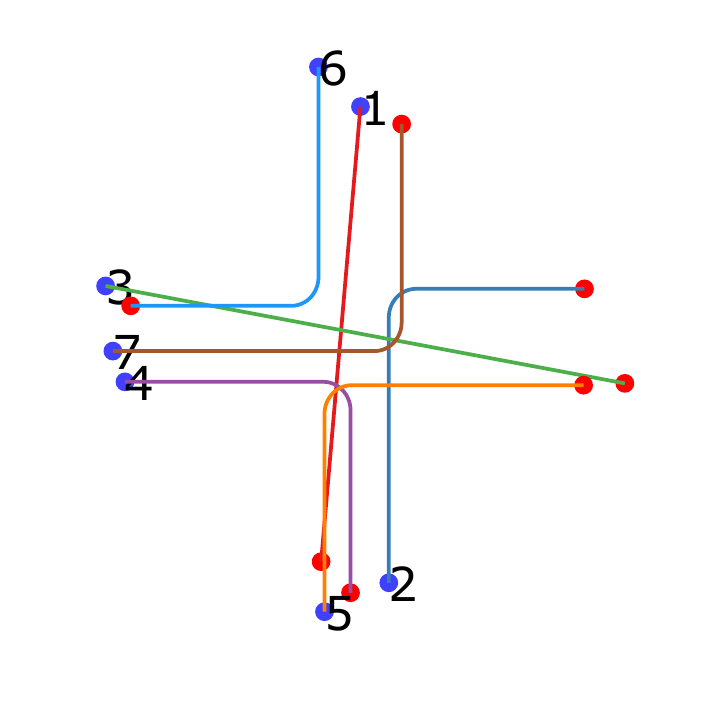}}{\scriptsize Instance 6, 7 robots}
    \\
    \stackon{\includegraphics[width=0.325\columnwidth]{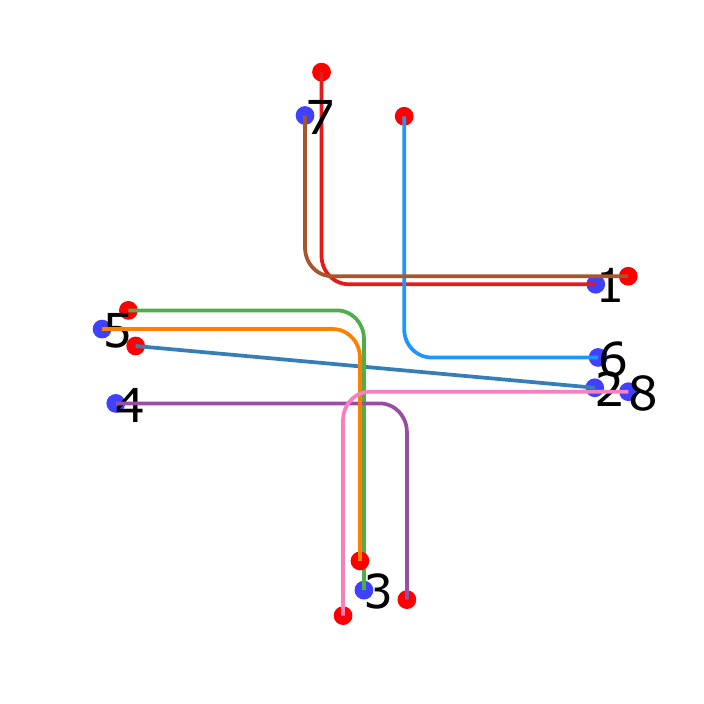}}{\scriptsize Instance 7, 8 robots}
    \stackon{\includegraphics[width=0.325\columnwidth]{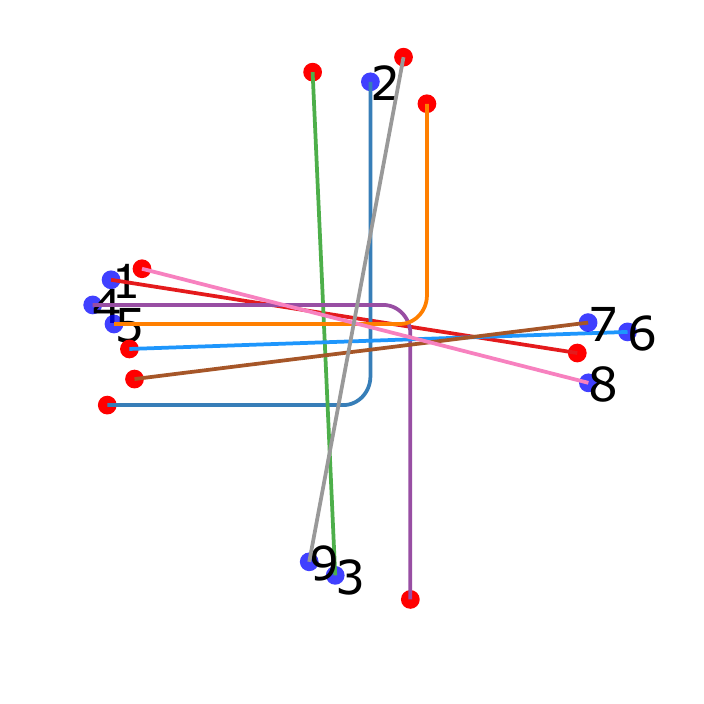}}{\scriptsize Instance 8, 9 robots}
    \stackon{\includegraphics[width=0.325\columnwidth]{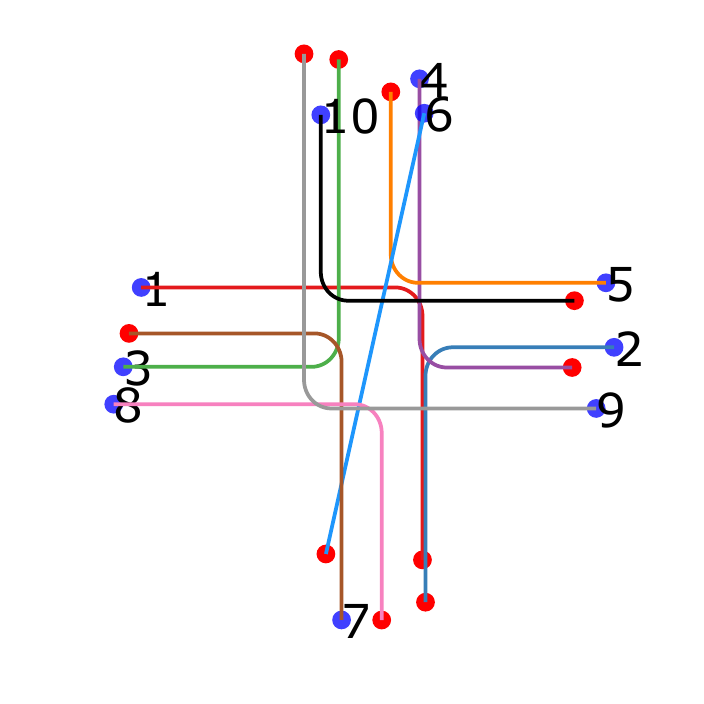}}{\scriptsize Instance 9, 10 robots}
    \\
\end{tblr}
\caption{Robot configurations in intersection environments and instances of the ideal/reference trajectories.}
\label{maps}
\end{figure}

\begin{figure}[t!]
\centering
\begin{tblr}{
    colspec = {Q[l]},
    rowsep = 1pt,
    colsep = 1pt,
    vlines = dashed,
    hlines = dashed,
    }
    \stackon{\includegraphics[width=0.32\columnwidth]{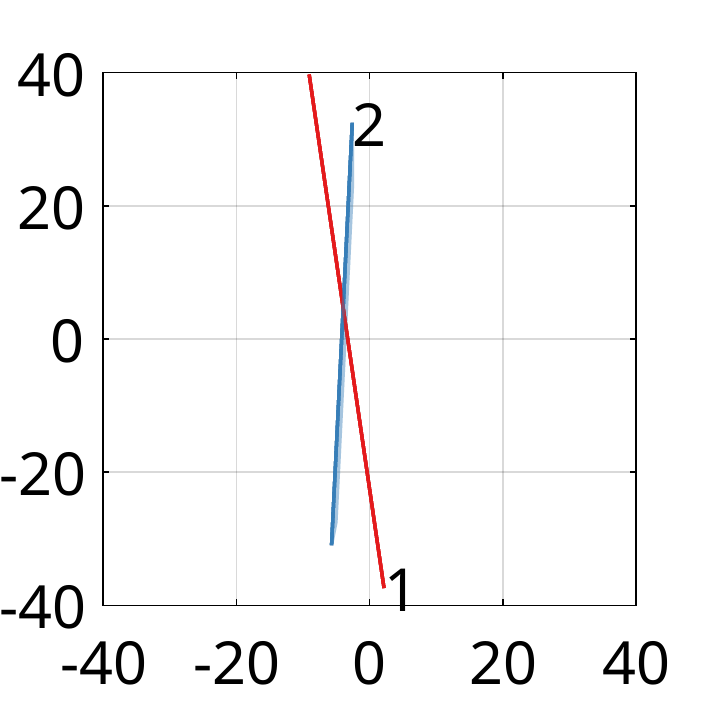}}{ Instance 1}
    \stackon{\includegraphics[width=0.32\columnwidth]{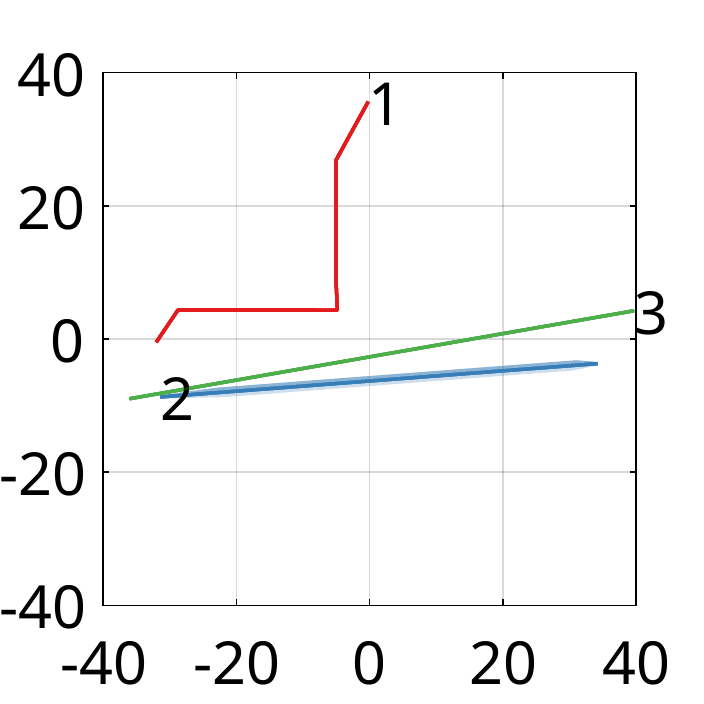}}{ Instance 2}
    \stackon{\includegraphics[width=0.32\columnwidth]{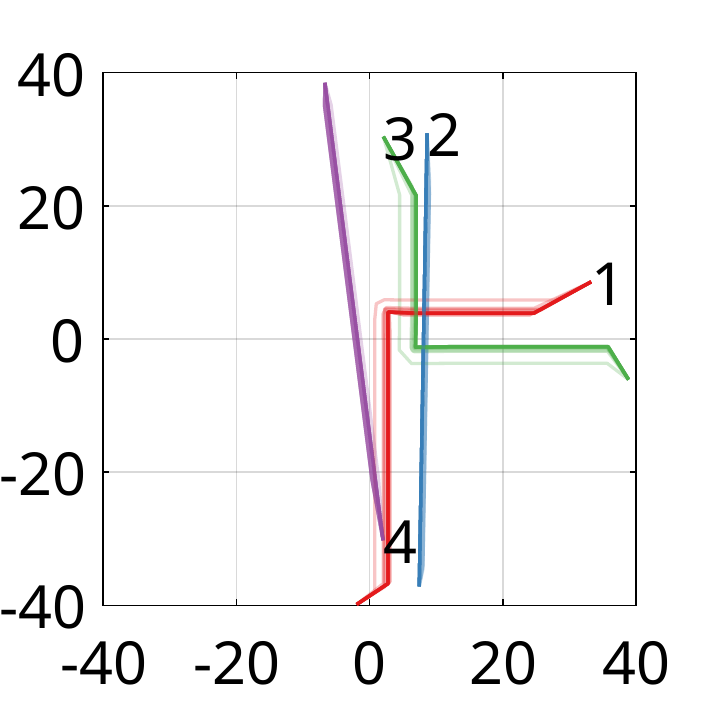}}{ Instance 3}
	\\
    \stackon{\includegraphics[width=0.32\columnwidth]{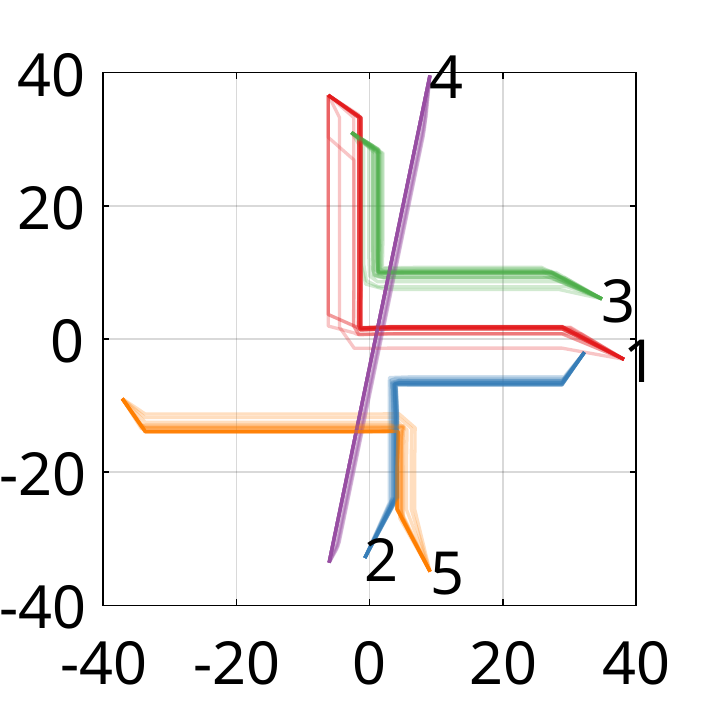}}{ Instance 4}
    \stackon{\includegraphics[width=0.32\columnwidth]{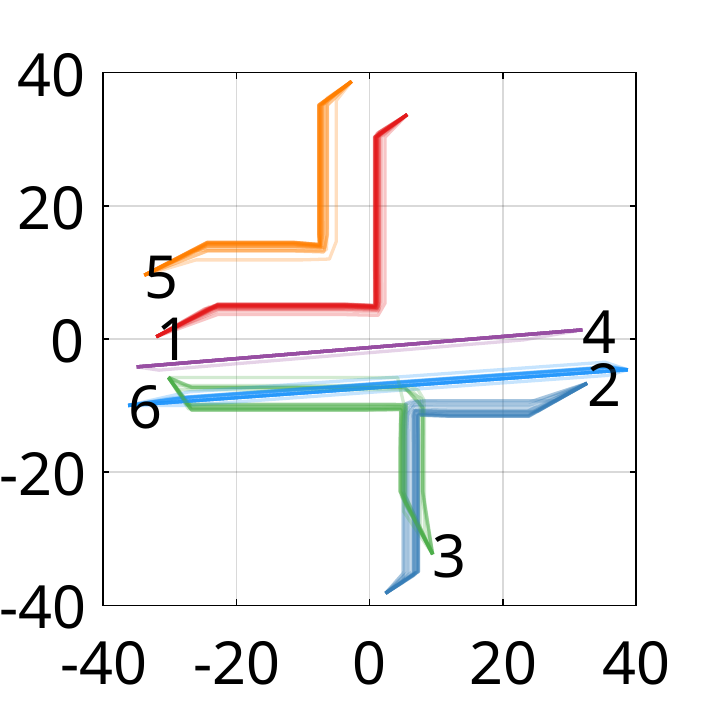}}{ Instance 5}
    \stackon{\includegraphics[width=0.32\columnwidth]{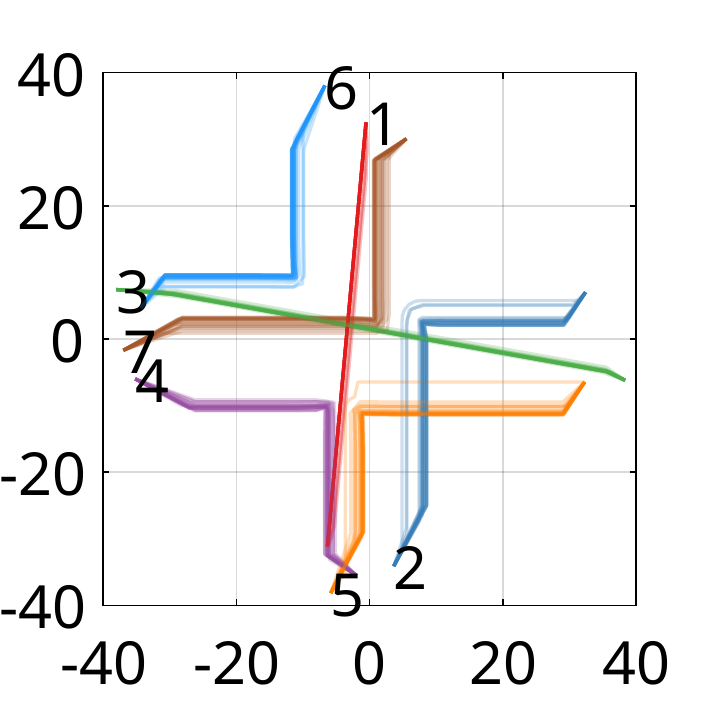}}{ Instance 6}
    \\
    \stackon{\includegraphics[width=0.32\columnwidth]{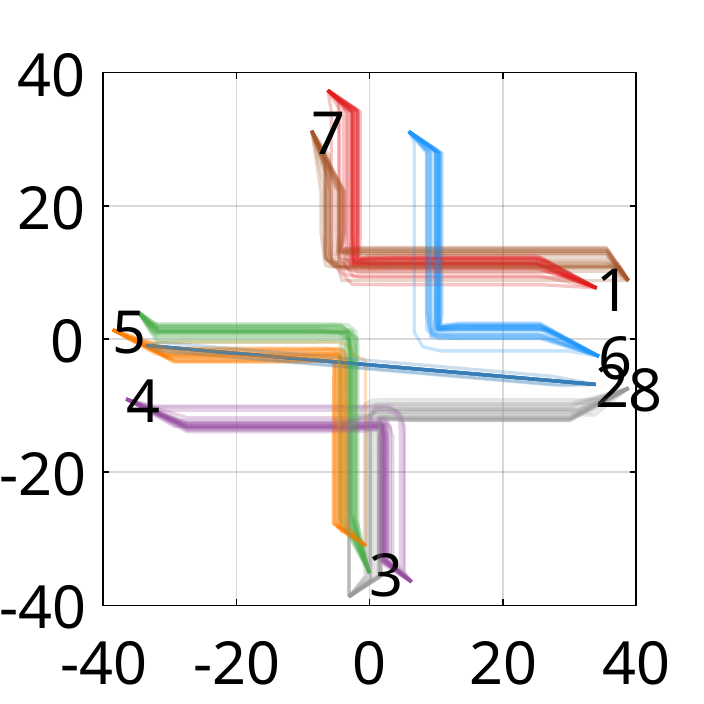}}{ Instance 7}
    \stackon{\includegraphics[width=0.32\columnwidth]{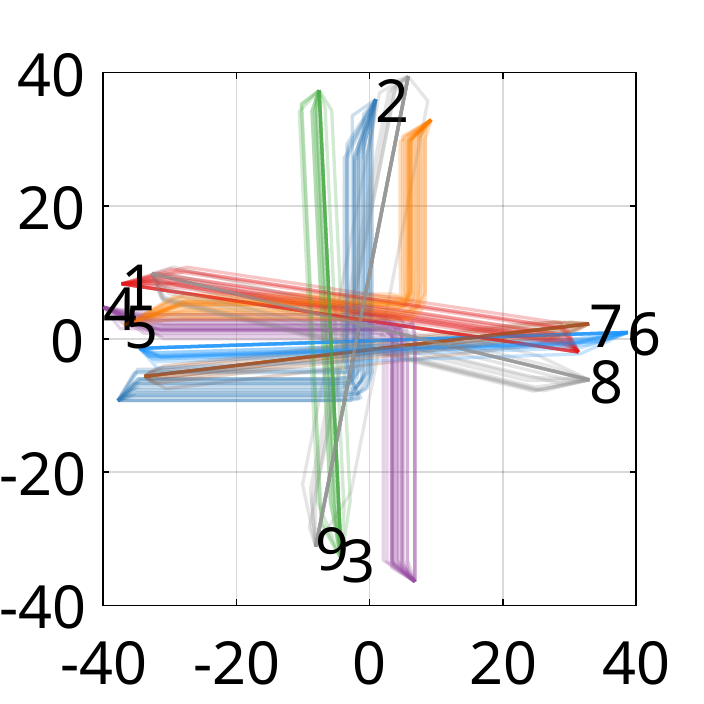}}{ Instance 8}
    \stackon{\includegraphics[width=0.32\columnwidth]{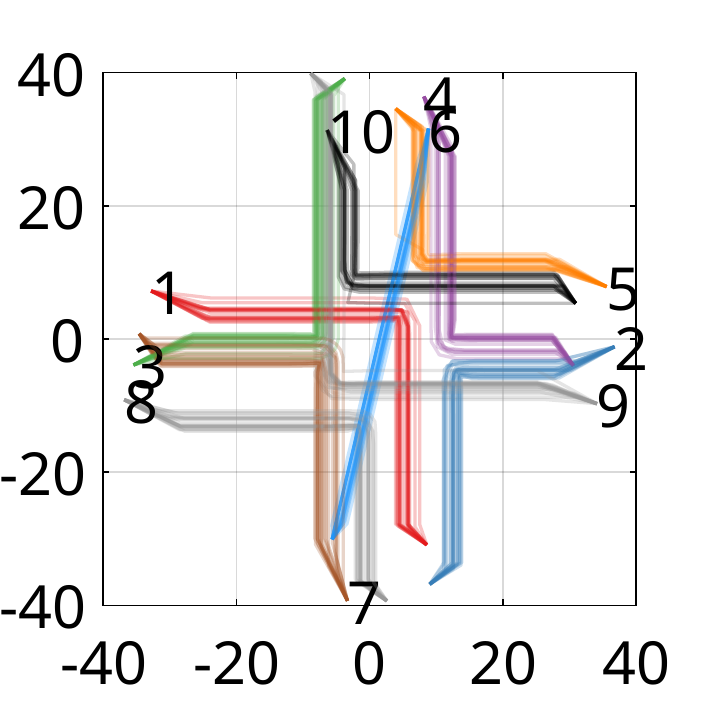}}{ Instance 9}
    \\
\end{tblr}
\caption{Multi-robot trajectories rendered by \algon{RADES}.}
\label{bmaps}
\end{figure}

\begin{figure*}[t!]
\centering
\begin{tblr}{
    colspec = {Q[l]},
    rowsep = 1pt,
    colsep = 0.1pt,
    vlines = dashed,
    hlines = dashed,
    }
    \stackon{\includegraphics[width=0.245\textwidth]{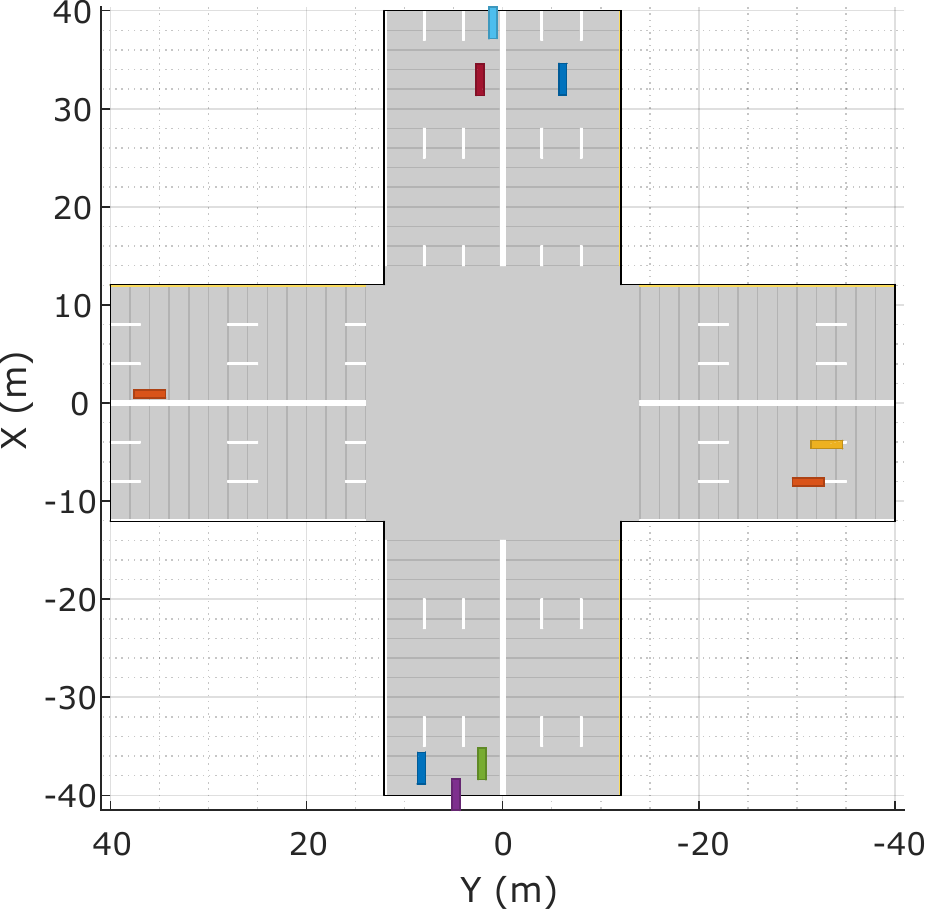}}{\scriptsize \textbf{Start Configuration: 1}}
    \stackon{\includegraphics[width=0.245\textwidth]{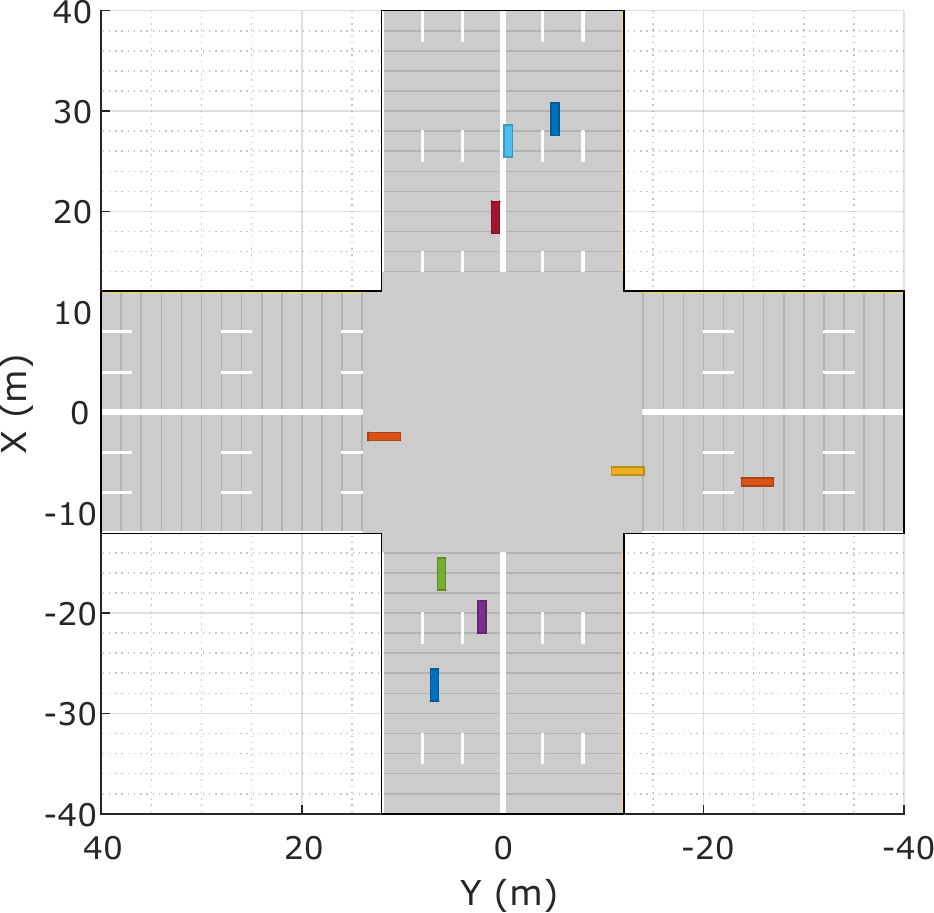}}{\scriptsize 2}
    \stackon{\includegraphics[width=0.245\textwidth]{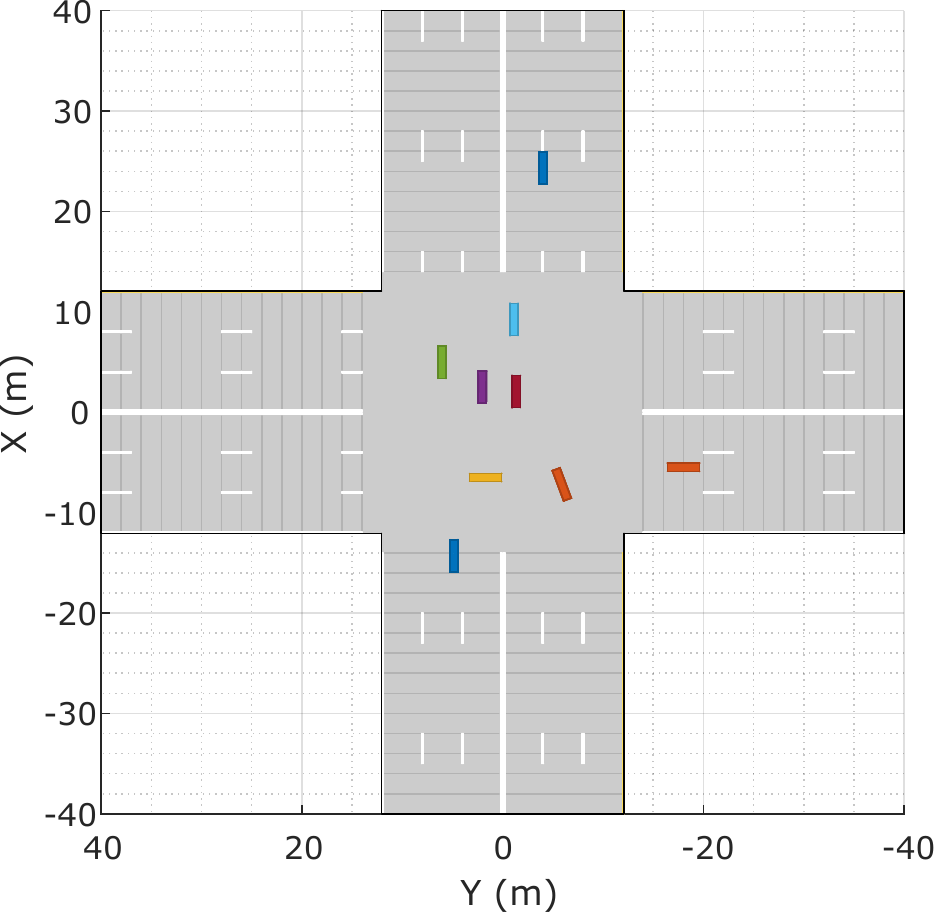}}{\scriptsize 3}
    \stackon{\includegraphics[width=0.245\textwidth]{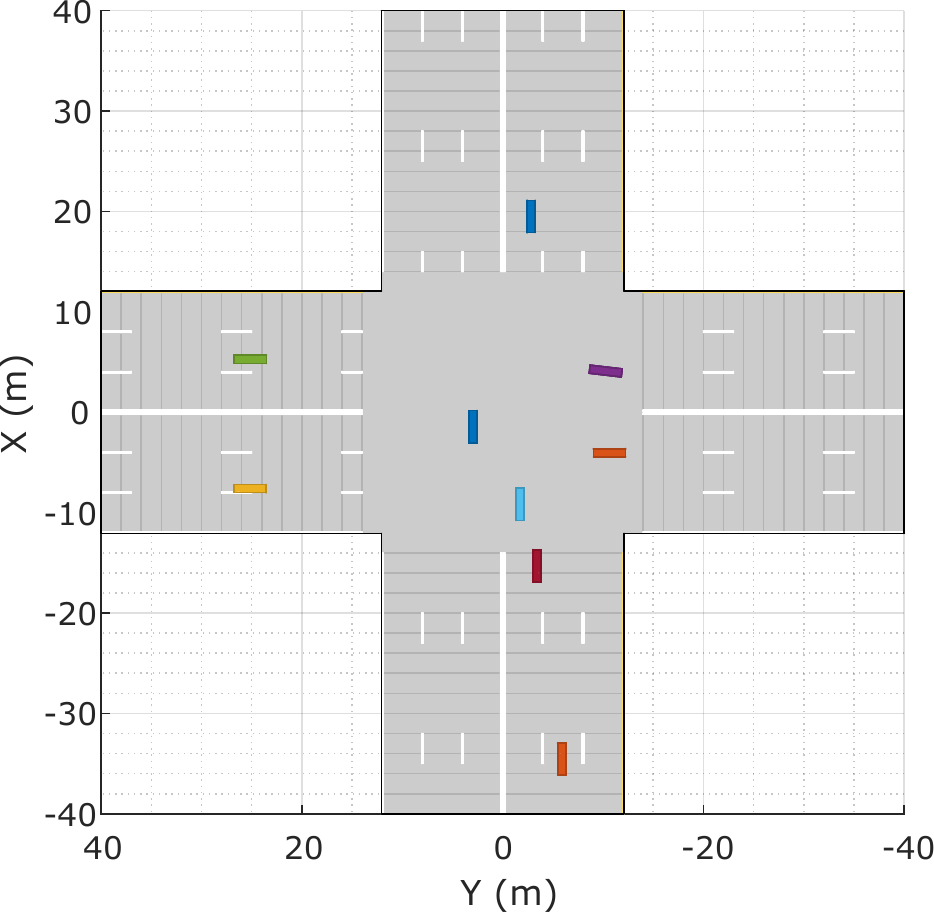}}{\scriptsize 4}
	\\
    \stackon{\includegraphics[width=0.245\textwidth]{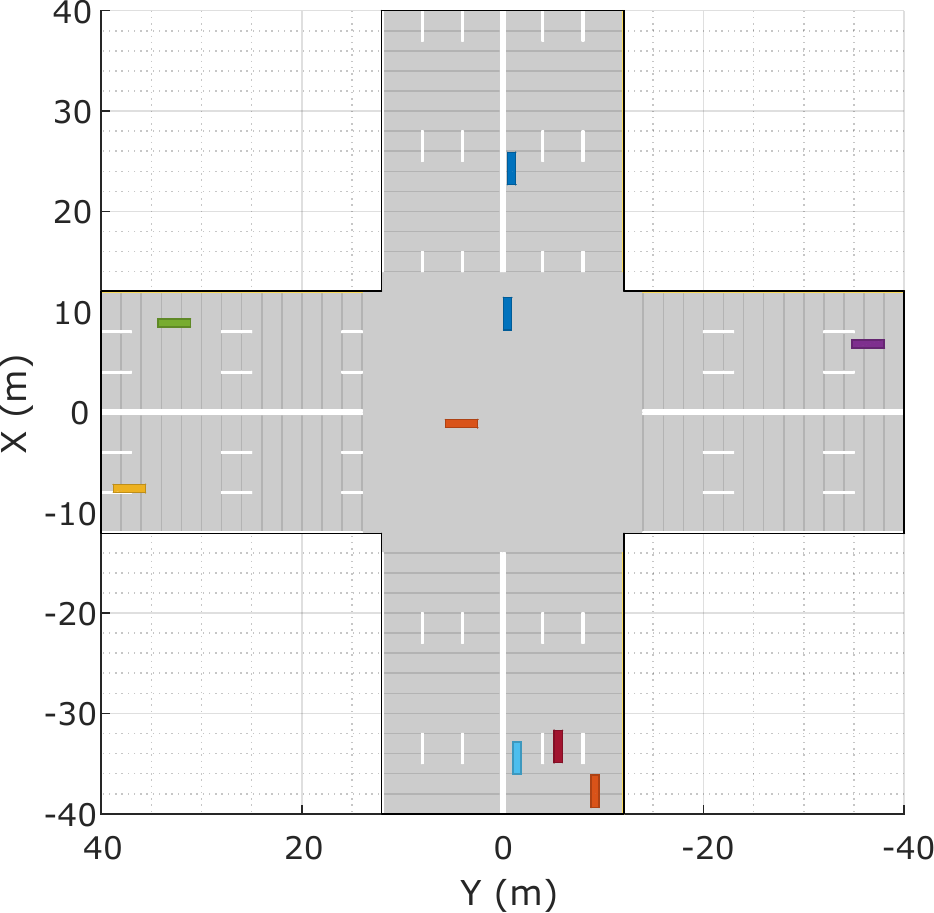}}{\scriptsize 5}
    \stackon{\includegraphics[width=0.245\textwidth]{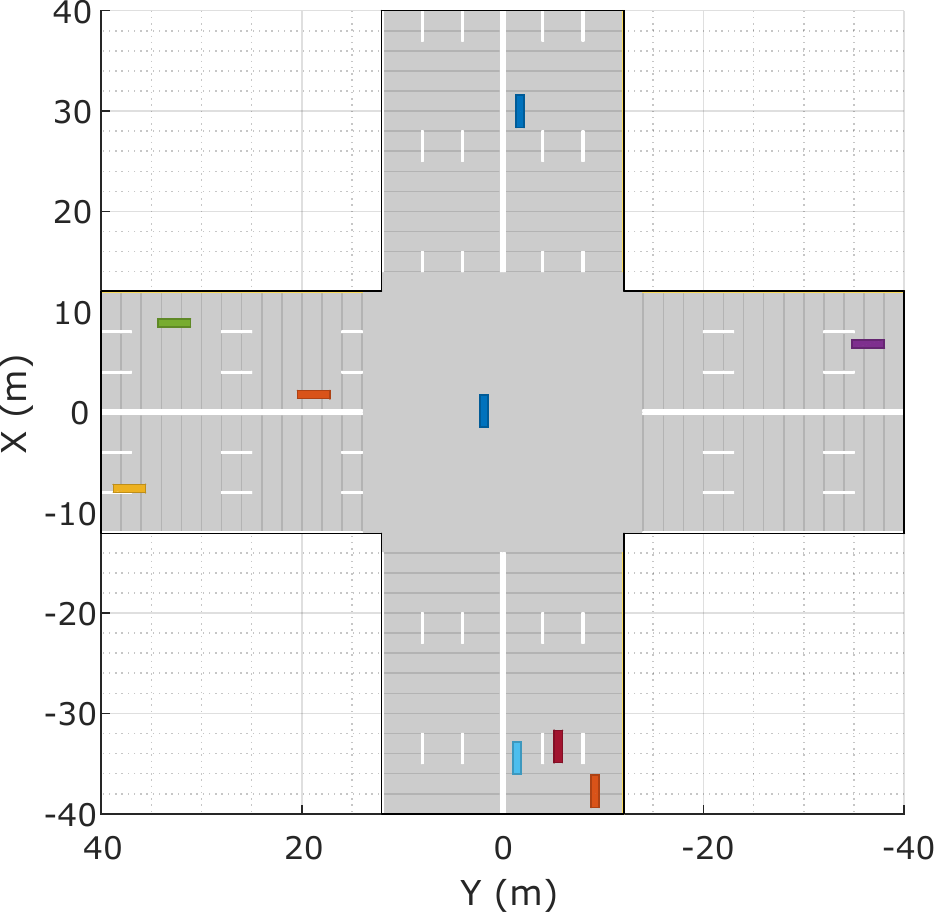}}{\scriptsize 6}
    \stackon{\includegraphics[width=0.245\textwidth]{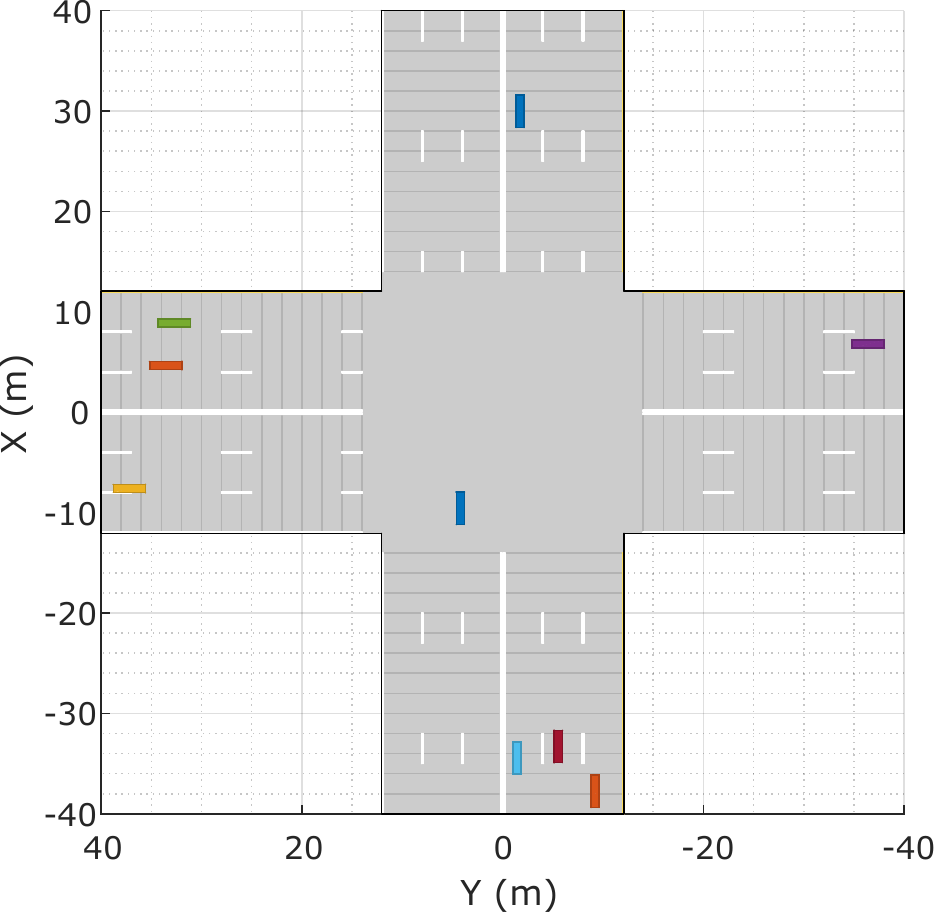}}{\scriptsize 7}
    \stackon{\includegraphics[width=0.245\textwidth]{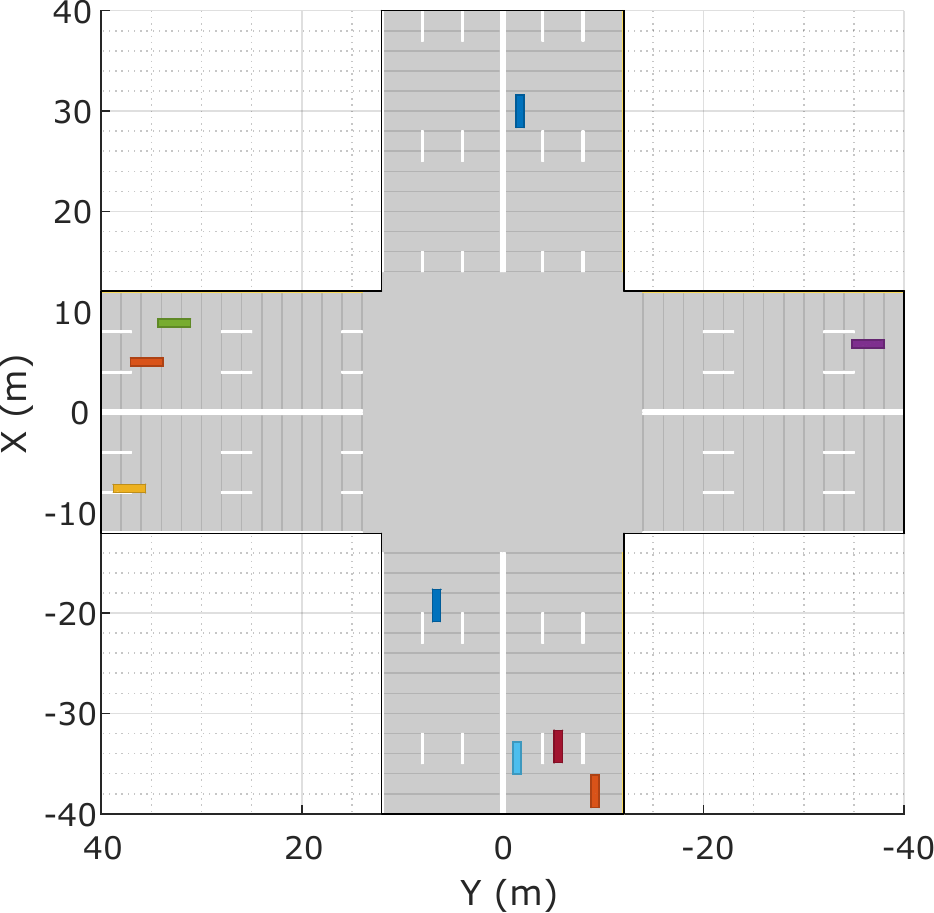}}{\scriptsize \textbf{End Configuration}}
    \\
\end{tblr}
\caption{Example of 9 robot navigation instances in an intersection scenario. The start configuration is portrayed in the top-left configuration, whereas the end configuration is portrayed in the bottom-right side. }
\label{example}
\end{figure*}

In order to show the characteristics of the rendered joint trajectories, Fig. \ref{maps} shows the configurations of the ideal trajectories for each scenario, and Fig. \ref{bmaps} shows the obtained collision-free multi-robot trajectories rendered by \algon{RADES} overall 20 independent runs. For ease of reference Fig. \ref{bmaps}, shows the trajectories of robots with different colors. One can observe from Fig. \ref{bmaps} that it is possible to obtain multiple feasible solutions with small number of function evaluations. Although the solutions with smaller number of robots leads to single joint trajectory solutions, the inclusion of larger number of robots induce in obtaining multiple solutions as shown by the non-overlapped trajectories in Fig. \ref{bmaps}. Although we used 300 function evaluations as an upper bound to allow the generation of feasible joint trajectory solutions, it is possible to use larger number of function evaluations to render trajectories with minimal length. 

Furthermore, since collision-free joint trajectories consist of time-parameterized waypoints, it is possible to render snapshots of the navigation of multiple robots in intersection domains. Fig. \ref{example} shows examples snapshots of the collision-free navigation of 9 robots. For simplicity and ease of reference, we show the transition over eight configurations out of a large number of simulation frames. As can be seen from Fig. \ref{example}, the robots are able to avoid collision during navigation. 

The observations derived from this study are useful to evaluate the convergence performance of the studied gradient-free optimization algorithms and to design efficient sampling schemes able to tackle the multi-coordination and trajectory planning of robots in constrained environments.

\section{Conclusion}

In this paper, we have studied the multi-robot coordinated planning at intersection scenarios and explored the feasibility of rendering collision-free trajectories using a new gradient-free optimization scheme implementing the difference of vectors and the rank-archive based mechanisms under low number of function evaluations. The proposed algorithm is labeled as the Rank-based Differential Evolution with a Successful Archive (\algon{RADES}), and embeds not only the rank-based selection mechanism of potential solutions, but also the archive to store successful mutations for subsequent sampling operations, as well as the stagnation counters to guide the sampling mechanisms.

Our computational experiments involving the multi-robot coordinated planning on ten types of intersection scenarios have shown the superiority of the proposed algorithm compared to seven other related optimization approaches. We also noted the attractive performance of archive and rank-based strategies to attain better convergence. On the other hand, algorithms using exploration strategies underperformed over a large number of cases. Overall instances, the results show the feasibility of generating multi-robot time-parameterized trajectories that avoid collision among entities.

Future venues for potential inquiry lie in our agenda. Investigating the role of parameter adaptation in optimization and evaluating other forms of graph structures for lattice roadmap configurations are left for future work. Also, investigating the scalability to larger map and number of robot instances, and developing the enhanced optimization heuristics are part of our agenda. Our results have the potential to elucidate new optimization-based approaches for multi-robot trajectory planning and navigation.

\bibliographystyle{IEEEtran}
\bibliography{mybiblio}

\end{document}